\documentclass[journal,lettersize]{IEEEtran} 
\usepackage{dblfloatfix}
\usepackage{moreverb,url}
\usepackage[T1]{fontenc}
\usepackage{float}
\usepackage{amsfonts}
\usepackage{amsthm}
\usepackage{pifont}
\usepackage{xcolor}
\usepackage{soul}

\newtheorem{definition}{Definition}
\newtheorem{proposition}{Proposition}

\newtheorem{problem}{Problem}
\usepackage{listings}
\usepackage{amsmath}
\usepackage{amssymb}
\usepackage{subfig}
\usepackage{graphicx}
\usepackage{hyperref}
\usepackage{csquotes}
\usepackage{tabularray}
\usepackage{varwidth}
\usepackage{booktabs}
\usepackage{makecell}
\newenvironment{notetopractitioners}%
{\small\textit{\textbf{Note to Practitioners}} --- \bfseries}%

\definecolor{SpringGreen}{RGB}{40,160,80} 
\definecolor{SkyBlue}{RGB}{60,130,200}   
\definecolor{Khaki}{RGB}{160,100,70}     
\definecolor{Lavender}{RGB}{180,100,180}  
\newcommand{\replytoall}[1]{\textcolor{SkyBlue}{#1}}
\newcommand{\replytoone}[1]{\textcolor{SpringGreen}{#1}}
\newcommand{\replytotwo}[1]{\textcolor{Khaki}{#1}}
\newcommand{\replytothree}[1]{\textcolor{Lavender}{#1}}

\newcommand{\deleted}[1]{}
\newcommand{\norm}[1]{\left\lVert#1\right\rVert}
\renewcommand{\replytoall}[1]{#1}
\renewcommand{\replytoone}[1]{#1}
\renewcommand{\replytotwo}[1]{#1}
\renewcommand{\replytothree}[1]{#1}
\newcommand{\replytotworoundtwo}[1]{\textcolor{Khaki}{#1}}
\newcommand{\replytoallroundtwo}[1]{\textcolor{SkyBlue}{#1}}
\renewcommand{\replytotworoundtwo}[1]{#1}
\renewcommand{\replytoallroundtwo}[1]{#1}

\usepackage{enumitem}
\setlist[itemize]{align=parleft,left=0pt..1em}

\begin{document}
\title{
Modular Multirotors: From Quadrotors to Fully-Actuated Aerial Vehicles
}

\author{
Jiawei Xu, Diego S. D'Antonio, and David Saldaña
\thanks{J. Xu, D. S. D'Antonio, and D. Salda\~{n}a are with the Autonomous and Intelligent Robotics Laboratory (AIRLab), Lehigh University, PA, USA:$\{$\texttt{jix519, diego.s.dantonio, saldana\}@lehigh.edu}}
}
\maketitle

\begin{abstract}
Traditional aerial vehicles are constrained to perform specific tasks due to their adhoc designs.
Based on modularity, we propose a versatile robot, H-ModQuad, that can adapt to different tasks by increasing its load capacity and actuated degrees of freedom.
It is composed of cuboid modules propelled by quadrotors with tilted rotors. We present two families of module designs that bring scalable and versatile actuation to the aerial systems. By configuring multiple modules, H-ModQuad can increase its payload capacity and change its actuated degrees of freedom from 4 to 5 and 6. 
By modeling the actuation capability of H-ModQuad using actuation ellipsoids and wrench polytopes, we find the body frame of a vehicle that maximizes its thrusting efficiency. We also compare the vehicle capabilities against formally defined task requirements. We present the dynamics of H-ModQuad and integrate control strategies despite the vehicle design. The design and model are validated with experiments using actual robots, showing that H-ModQuad vehicles with different configurations provide different actuation properties.
\end{abstract}

\begin{notetopractitioners}
This paper presents the H-ModQuad, a modular multirotor system designed to enhance the flexibility and versatility of aerial vehicles. Traditional drones often need to be specifically designed or modified for different tasks, but H-ModQuad allows for quick reconfiguration to meet varying operational needs. By adding or rearranging modules, users can easily increase the vehicle's load capacity or improve its force and torque generation characteristics. This adaptability makes H-ModQuad particularly useful in industries such as logistics for transporting varying payloads, inspection and maintenance tasks in hard-to-reach areas, and unmanned tool wielding. Our analytical tools provide pre-assembly vehicle capability assessment, and flight tests demonstrate that different modular configurations grant the vehicle with distinct actuation characteristics suitable for diverse tasks without redesigning the modules. Therefore, H-ModQuad offers a versatile aerial solution, making it a valuable tool for a wide range of autonomous applications.
\end{notetopractitioners}


\section{Introduction}
Research on unmanned aerial vehicles (UAVs) has thrived in recent years, 
offering low-cost solutions for a wide spectrum of applications such as aerial surveillance~\cite{9797309}, transportation~\cite{pei2022urban}, and construction~\cite{10021301}.
However, a major problem with these vehicles is the lack of versatility. There are two characteristics of versatility on which we would like to focus, the \emph{strength} and the \emph{actuated degrees of freedom (ADoF)}.
For instance, in object transportation, the hardware design depends on the weight and dimensions of the target payload. An underpowered UAV cannot transport a heavy payload in time and safety. Although an oversized UAV can carry both large and small objects, it suffers from higher energy consumption and limited motion capability in cluttered environments. 
Having additional \emph{ADoF} is the other desirable property for UAVs. For instance, a traditional quadrotor has four co-linear rotors, making the vehicle under-actuated. Existing solutions choose to control the three positional DOF and one DOF in the yaw angle of a quadrotor~\cite{5717652}, requiring to tilt for horizontal translation.
Adding redundant vertical rotors to a quadrotor~\cite{alaimo2013mathematical}, or connecting multiple modular quadrotors without changing the rotor orientation~\cite{8461014}, does not increase the ADoF. As long as all the rotors point in the same direction, it remains impossible for the vehicles to translate horizontally without tilting.

\begin{figure}[t!]
    \centering
    \includegraphics[width=\linewidth]{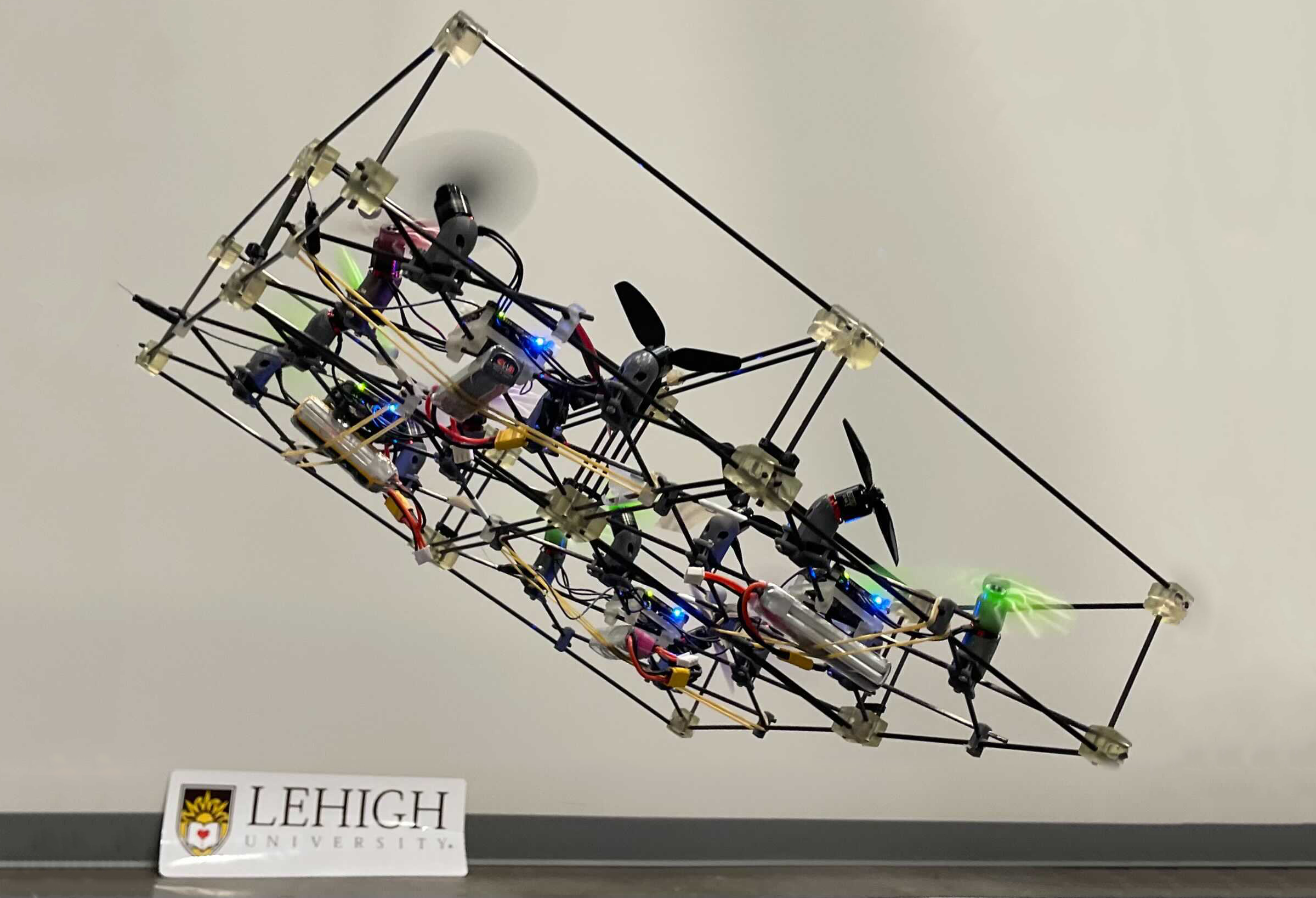}
    \caption{Four heterogeneous torque-balanced modules of two types forming an 2-by-2 H-ModQuad structure. Each type of modules are diagonally placed in the structure. This structure is used in \replytotworoundtwo{Experiment 4}, and the picture is captured during its flight. The video of all narrated validations can be found at~\url{https://youtu.be/Mxo0QpxU-9A}.}
    \label{fig:titlePic}
\end{figure}

To address the limitation in underactuation, an approach is to increase the number of rotors in a multirotor vehicle and change their orientation~\cite{7139759}. The orientation of the rotor can be fixed or actively changed. The problem with actively tilting rotors is that the tilt mechanism increases the complexity of vehicle and controller design, such as the articulated aerial robot~\cite{8258850}.
The work in~\cite{franchi2018full} provides a universal control strategy for both underactuated and fully-actuated UAVs, but the design of the UAV is specific for certain tasks, which limits the versatility of each individual UAV design.

Modular aerial robots offer a versatile and scalable way to increase their strength~\cite{8461014,duffy2015lift} and capabilities~\cite{yang-icra19-aerialskeleton,zhaoIJRR2019,modquadgripper}. When the task changes, a modular robot can adapt to new requirements by either adding modules or reconfiguring the modules. In addition to increasing the maximum force output, having more modules also provides redundancy to the system, which leads to enhanced robustness against mechanical deficiency or noise. 
\replytotwo{As highlighted in the literature, modular robots can adapt to new tasks by swapping modules depending on the task without the need of redesigning the entire system~\cite{gilpin2010modular,9507292,9842365}. With special designs, a modular system can even obtain new structural characteristics beyond a simple scaling to the quantity of modules~\cite{9410285}.}
To tackle transportation tasks, a modular UAV can change the number of modules in the vehicle to adapt to different payloads, for example, a pizza box versus a piano. For maintenance tasks such as drilling, the vehicle would reconfigure to enhance its actuation for independent translation, \textit{i.e.},
although the ADoF remains the same after reconfiguration, the maximum force that the vehicle can generate in a selected direction increases.

\noindent
\textbf{Contribution:}
The main contribution of this paper is twofold.

\replytoall{1) We formally introduce the design concept of torque-balanced quadrotor modules that generate zero torque when all rotors produce identical thrust forces. Based on this concept, we propose two families of quadrotor modules, $R$- and $T$-modules, and prove that both families satisfy the torque-balanced property.}
\replytoall{2) We propose a model based on polytopes in the $6$-D wrench space to formalize the concept of vehicle actuation capabilities.}

\replytotwo{While similar torque-balanced module designs exist in the literature~\cite{8453359,allenspach2020design,7989608}, our design enables H-ModQuad vehicles to achieve unique actuation capabilities, such as full actuation, through different modular assemblies. Compared to force envelopes~\cite{8206269,franchi2018full,hamandi2021design}, our actuation polytopes account for realistic constraints, including motor saturation and force-torque coupling. With formally defined actuation requirements for tasks as sets of $6$-D wrenches, we provide methods to evaluate an H-ModQuad vehicle's polytope against task requirements to verify its suitability.}

\replytotwo{Our previous work~\cite{9561016} introduced $R$-modules and H-ModQuad vehicles with varying ADoF based on heterogeneous module configurations. Generalizing $R$-modules to the torque-balanced module concept, this paper introduces \textbf{a new family of quadrotor designs, called $T$-modules}, which feature enhanced torque generation. Using the \textbf{newly developed actuation capability model}, we demonstrate that a fully-actuated vehicle assembled solely of $T$-modules can hover at significantly higher tilt angles than one composed only of $R$-modules. \textbf{Three new experiments} validate this theory, complement the original experiments, and highlight the increasing ADoF of H-ModQuad vehicles composed of torque-balanced modules.
}

\section{Related Work}
\begin{figure}[t]
    \centering
    \includegraphics[width=\linewidth]{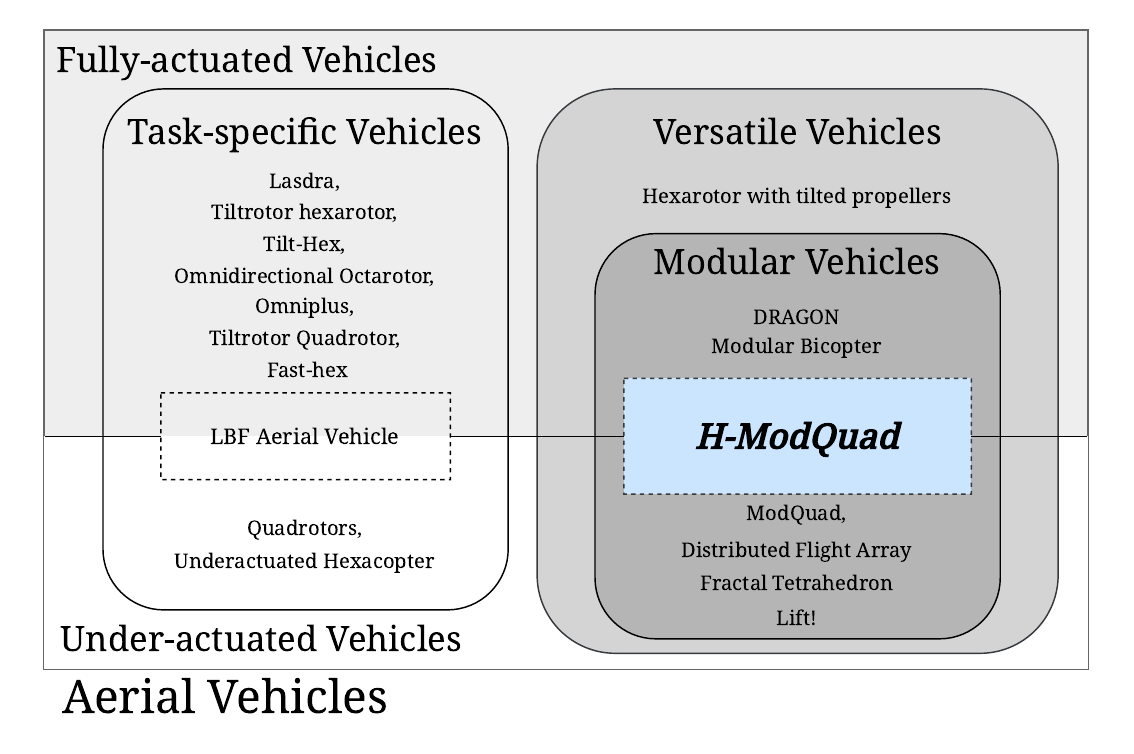}
    \caption{The taxonomy diagram of all aerial vehicle designs mentioned in this paper. We consider the their characteristics of versatility, modularity, and ADoF. Note that our modular multirotor vehicle design finds its classification in the shaded cell as \emph{a modular versatile design that can change its ADoF}. }
    \label{fig:taxonomy}
\end{figure}

We review multirotor designs related to the versatility problem in the literature. A taxonomy diagram based on whether designs can adapt to different tasks, their modularity, and their ADoF, is shown in Fig.~\ref{fig:taxonomy}.

A traditional quadrotor belongs to the category of task-specific underactuated vehicles, limited by its payload capacity and ADoF. One method of increasing its payload limit is to increase the number of rotors in the vehicle, such as a hexarotor~\cite{alaimo2013mathematical}, although the co-linear rotors indicate the design is still underactuated.
The work in~\cite{7139759} introduces the design and control of a hexarotor with tiltable rotors. Using six unidirectional rotors, the vehicle can actuate up to 6~degrees of freedom. By configuring the rotor angles, the robot minimizes energy consumption when following a certain trajectory. However, trajectory-specific optimization limits the versatility of this approach. By integrating eight bidirectional motors, the authors in~\cite{7487497} present a design that achieves omnidirectional motion. This design enables the multirotor vehicle to hover in an arbitrary attitude at the cost of structural complexity and energy consumption. The work in~\cite{8281444, 9214065} presents an omnidirectional multi-rotor design with no less than seven unidirectional motors.~\cite{8281444} show a controller that guarantees nonnegative thrust forces for all rotors, ensuring that the desired states are achievable using unidirectional motors.~\cite{9214065} provide an optimization of motor placement to minimize control input.

As an alternative to increasing the number of rotors,~\cite{6225129} presents the controller for a quadrotor with actively tilting rotors to achieve full actuation. 
~\cite{7759271} presents a hexarotor that uses a servomotor to tilt the rotors synchronously. The ability to transition between under actuation and full actuation allows the vehicle to choose operating modes between high efficiency and full actuation. 
These vehicle designs feature fixed mechanisms. Given a new task beyond their original capabilities, the multirotors cannot meet the requirements.

\replytotwo{We consider modularity a key characteristic for a robot design to be versatile in different tasks, as validated in ground vehicles~\cite{9991984,9115061}, water-surface vehicles~\cite{10474289}, and manipulator arms~\cite{9962372}. Researchers apply similar principles to multirotor vehicles.} The works in~\cite{5509882,doi:10.1177/0278364913501212} present a hexagonal one-rotor module that composes a flight array with increasing strengths.~\cite{9172614} develop a modular multirotor system based on elementary modules of tetrahedron-shaped quadrotors.~\cite{8461014} introduce a cuboid quadrotor module that flies independently and assembles with other modules in-air into a larger structure. Adding more modules to a structure improves redundancy and payload capacity, but the ADoF remains the same since all rotor forces are parallel. The quadrotor modules presented in~\cite{10563193} use micro-servos to actively tilt the rotors, achieving additional ADoF and enabling full actuation.~\cite{9982090} show that connecting bicopter modules equipped with actively tilting rotors increases load capacity and ADoF, allowing the assembled multirotor vehicle to meet diverse task requirements. \replytotwo{Modular multirotors typically require special controllers that distribute actuation across modules while handling disturbances caused by modular connections. Researchers have explored approaches such as modular reinforcement learning~\cite{10777540}, robust control~\cite{robotics10020076}, and distributed control~\cite{10606878}.} These modular vehicle designs adapt to different tasks, such as transporting heavier payloads and following higher-DOF trajectories, making them versatile solutions.

Among the existing literature, researchers are striving to increase the actuation capabilities and versatility of multirotor vehicles by deploying novel designs, applying novel control strategies, and utilizing modularity. However, few designs have considered the use of modular robots that adapt to various tasks without reconstructing modules. In this work, we seek to fill this gap by presenting a versatile modular multirotor vehicle composed of lightweight modules that can achieve a varying number of ADoF that can be applied to perform a variety of task with different requirements.

\section{Problem Statement}
This work focuses on modular multirotor vehicles composed of $n$ modules, denoted by the set $\mathcal{M} = \{M_1, M_2, \cdots, M_n\}$.
\begin{definition}[Module]
A \emph{module} is an autonomous quadrotor within a cuboid frame.
The rotors do not necessarily have to be vertical to the bottom face of the module, and their orientation defines the actuation characteristics of the module. 
\end{definition}

Module $M_i\in\mathcal{M}$ has a mass~$m_i$ and an inertia tensor~$\boldsymbol{J}_i$.
Multiple modules can connect to each other by aligning their sides to
create a rigid connection, which assembles a \emph{structure}.
\begin{definition}[Structure]
A \emph{structure} is a set of $n\geq1$ rigidly connected modules
forming a single multirotor vehicle. 
Its inertia tensor is denoted by $\boldsymbol{J}$ and its mass by $m$.
\end{definition}

\begin{figure}[t]
\centering
    \includegraphics[width = 0.2\linewidth]{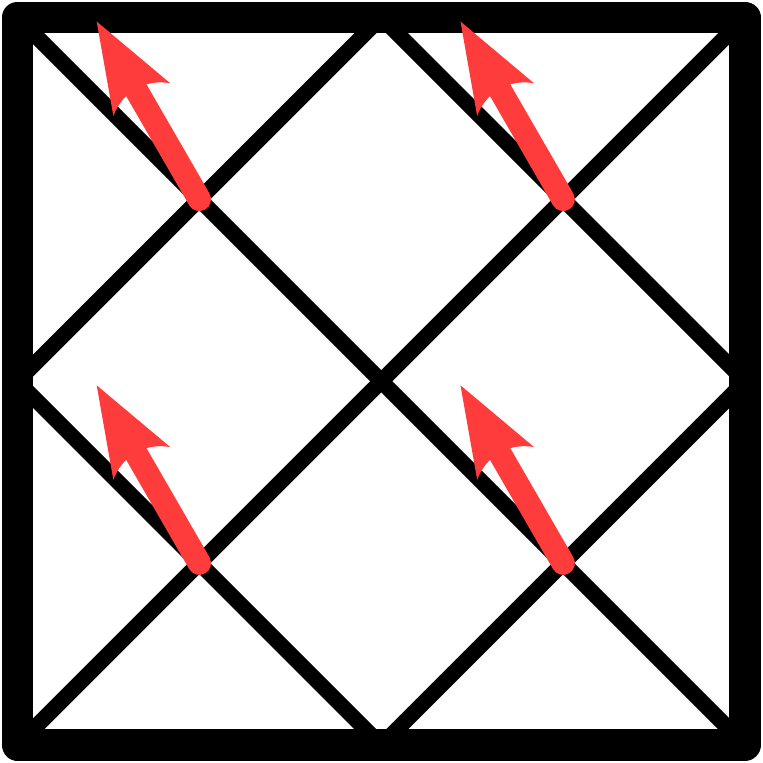}\quad\includegraphics[width=0.7\linewidth]{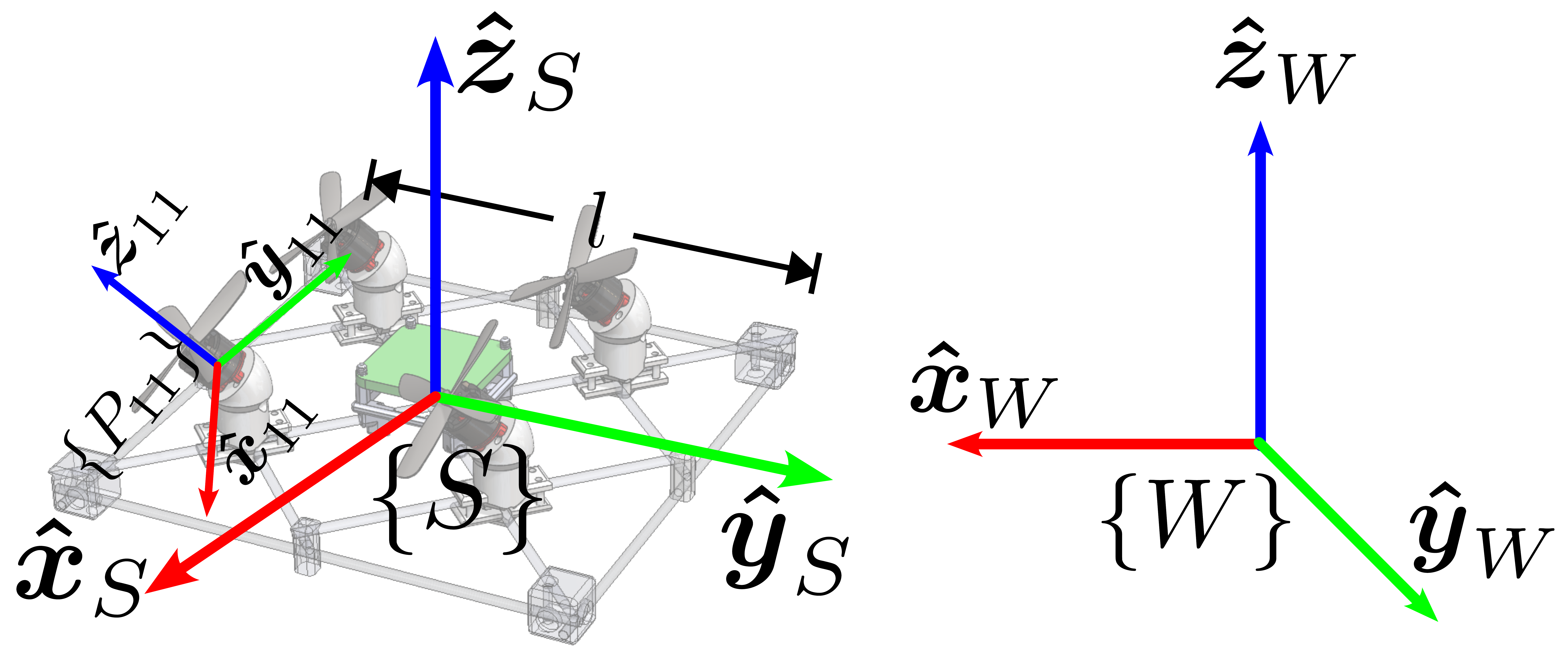}
    \caption{An $R$-module with its coordinate frames, and dimensions. Note that this module composes a structure, thus $\{S\}$ aligns with $\{M_1\}$.}
    \label{fig:oneUAV}
\end{figure}

We denote the standard basis in $\mathbb{R}^3$ by $\boldsymbol{\hat x}=\left[1,0,0\right]^\top,\: \boldsymbol{\hat y}=\left[0,1,0\right]^\top,$ and $\boldsymbol{\hat z}=\left[0,0,1\right]^\top$. 
The world reference frame $\{W\}$ is fixed with its $z$ axis pointing upward. 
Module $i$ in the structure has a module frame $\{M_i\}$ with its origin in the module's center of mass (COM). We define the ``front'' direction of the module as the $x$-axis and the ``up'' as the $z$-axis.
The four rotors are located on the $xy$-plane of $\{M_i\}$ in a square configuration. Each rotor has a rotor frame, $\{P_{ij}\}$, with its $z$-axis pointing in the direction of the rotor force. The orientation of $\{P_{ij}\}$ in $\{M_i\}$ is specified by the rotation matrix ${}^{i}\boldsymbol{R}_{j}\in\mathsf{SO}(3)$. 
In a traditional quadrotor, all the rotors are parallel to the $z$-axis of the quadrotor. In our case, the rotors can point in different directions. The associated coordinate frames of a module are illustrated in Fig.~\ref{fig:oneUAV}. The structure frame, denoted by $\{S\}$, has its origin in its COM. Without loss of generality, we align all module frames in the structure and define the $x$-, $y$- and $z$-axes of $\{S\}$ as in parallel to the $x$-, $y$- and $z$-axes of all modules in the structure. 
The location and orientation of $\{S\}$ in the world frame $\{W\}$ is specified by the vector $\boldsymbol{r}\in\mathbb{R}^3$ and the rotation matrix ${}^{W}\!\!\boldsymbol{R}_{S}\in \mathsf{SO}(3)$.

The spinning of rotor $j\in\{1,...,4\}$ in module $i$ 
generates a thrust $f_{ij}\in\left[0, f_{max}\right]$,
and an air drag. The resulted force and torque in $\{M_i\}$ are 
$
\boldsymbol{f}_{ij}=f_{ij}\:  {{}^{i}\boldsymbol{R}_{j} \boldsymbol{\hat z}}  
\text{, and }    
\boldsymbol\tau_{ij}=f_{ij}\: {(-1)^{j}\frac{k_m}{k_f}\:   {}^{i}\!\boldsymbol{R}_{j}\boldsymbol{\hat z}},
$
where $k_f$ and $k_m$ are coefficients experimentally obtained.
The structure generates total force~$\boldsymbol{f}$ and torque~$\boldsymbol\tau$ in the structure frame~$\{S\}$ which is the vector summation of all rotor forces and torques,
\begin{equation}
    \boldsymbol{f} = \sum_{ij}{}^{S}\!\boldsymbol{R}_{i}\boldsymbol{f}_{ij},\quad \boldsymbol\tau = \sum_{ij}\boldsymbol{\tau}^{f}_{ij}+\boldsymbol{\tau}^{d}_{ij},
    \label{eq:ftaustructure}
\end{equation}
where $\boldsymbol{\tau}^{f}_{ij} = \boldsymbol{p}_{ij}\times{}^{S}\!\boldsymbol{R}_{i}\boldsymbol{f}_{ij}$ is the torque generated by the thrust and $\boldsymbol{\tau}^{d}_{ij}={}^{S}\!\boldsymbol{R}_{i}\boldsymbol\tau_{ij}$ is the air drag; $\boldsymbol{p}_{ij}\in\mathbb{R}^3$ is the position of each rotor in $\{S\}$, and ${}^{S}\!\boldsymbol{R}_{i}$ is the orientation of module $i$ in $\{S\}$. Letting $\boldsymbol{u} = \left[f_{11}, f_{12}, ...,  f_{n4}\right]^{\top}$ be the input vector, we combine \eqref{eq:ftaustructure} in a matrix form to define the $6\times1$ wrench
{\footnotesize
\begin{eqnarray}
    \boldsymbol{w} &=& \boldsymbol{A}\boldsymbol{u},\qquad\text{where}\label{eq:MAu}\\
    \boldsymbol{A} &= \begin{bmatrix}
        \boldsymbol{A_f}\\
        \boldsymbol{A_\tau}
    \end{bmatrix} =& \begin{bmatrix}
    \dots&\!\!
    {}^{S}\!\boldsymbol{R}_{ij}\boldsymbol{\hat z}&\!\!\cdots \\
    \dots&\!\! \boldsymbol{p}_{ij}\!\!\times\!\!{}^S\!\boldsymbol{R}_{ij}\boldsymbol{\hat z}\!\!+\!\!{}^{S}\!\boldsymbol{R}_{ij}(-1)^{i+j}\frac{k_m}{k_f}\boldsymbol{\hat z} &\!\!\cdots
    \end{bmatrix}
    \label{eq:Acomponents}
\end{eqnarray}}
\noindent
is a $6\times4n$ \emph{design matrix} that maps the input forces into the total wrench in $\{S\}$; and ${}^{S}\!\boldsymbol{R}_{ij} = {}^{S}\!\boldsymbol{R}_{i}{}^{i}\boldsymbol{R}_{j}$. The design matrix is also known as the \emph{allocation} matrix~\cite{9214065}.
In traditional multirotor vehicles, this matrix is fixed after building the robot. In our case, this matrix can change depending on the module configuration.
The dynamics of the structure can be described using the Lagrangian for robot motion~\cite{Murray1994AMI,194594} as
\begin{equation}
    \boldsymbol{M}\begin{bmatrix}
        \ddot{\boldsymbol{r}}\\
        \dot{\boldsymbol\omega}
    \end{bmatrix} + \boldsymbol{C}\begin{bmatrix}
        \dot{\boldsymbol{r}}\\
        \boldsymbol\omega
    \end{bmatrix} + \boldsymbol{g} = \boldsymbol{Bw},
    \label{eq:Lagrange}
\end{equation}
where
$    
\boldsymbol{M} = \begin{bmatrix}
    m\boldsymbol{I}_3\! &\! \boldsymbol{0}\\
    \boldsymbol{0}\! &\! \boldsymbol{J}
\end{bmatrix},
\boldsymbol{C} = \begin{bmatrix}
    \boldsymbol{0}\! &\! \boldsymbol{0}\\
    \boldsymbol{0}\! &\! \boldsymbol{\omega}^\times\boldsymbol{J}
\end{bmatrix},
\boldsymbol{B} = \begin{bmatrix}
    {}^W\!\!\boldsymbol{R}_S\! &\! \boldsymbol{0}\\
    \boldsymbol{0}\! &\! \boldsymbol{I}_3
\end{bmatrix}
$
are the mass matrix, the Coriolis matrix, and the wrench transformation from the structure's local frame to the world frame, respectively. $\boldsymbol{\dot{r}}, \boldsymbol{\Ddot{r}}$ are the linear velocity and acceleration of the structure; $\boldsymbol\omega, \boldsymbol{\dot{\omega}}$ are the angular velocity and acceleration of the structure in $\{S\}$, respectively. The $(\cdot)^\times$ operator converts a vector into the corresponding skew-symmetric matrix. $\boldsymbol{g} = [mg\boldsymbol{\hat z}^\top\quad \boldsymbol{0}^\top]^\top$ is the gravity, $g$ is the gravitational acceleration, and $\boldsymbol{I}_3$ is the $3\times3$ identity matrix.

Note that $\text{rank}(\boldsymbol{BA})$
defines the number of actuated degrees of freedom. Regarding~\eqref{eq:Lagrange}, it determines the number of entries in $[\ddot{\boldsymbol{r}}^\top\quad\dot{\boldsymbol\omega}^\top]^\top$ that the vehicle can control independently. 
Since $\boldsymbol{B}$ is composed of a rotation matrix and an identity matrix diagonally, it is always full-rank. Thus, the ADoF depend on $\text{rank}(\boldsymbol{A})$. 

We design the modules with fixed rotor configurations that fly independently. When assembled, they can 
have 4, 5, or 6 ADoF.
The modular configuration determines the actuation capabilities of a structure, such as its ADoF, maximum tilt angle, and maximum strength. Since the design matrix $\boldsymbol{A}$ of a structure incorporates the rotor configuration information, we model the actuation capabilities of a structure based on $\boldsymbol{A}$.
We implement a controller for the multirotors that considers any structure regardless of its ADoF or module configuration. 
In this paper, we focus on solving the following problems.

\begin{problem}[Module Design] Given a desired direction of the maximum thrust of a module, specified by the rotation matrix $\boldsymbol{R}^\star$, design a module that can hover while maintaining the given orientation.
\label{p:moduledesign}
\end{problem}

\begin{problem}[Modeling Actuation Capabilities] Given a structure with design matrix $\boldsymbol{A}$, develop an analytical model that formalizes the robot capabilities.
\label{p:actuation}
\end{problem}

\begin{problem}[Controlling H-ModQuad]
Given a structure with either 4, 5 or 6 actuated degrees of freedom, 
derive a trajectory tracking controller that combines the three cases.
\label{p:nonomnicontrol}
\end{problem}

\section{Torque-Balanced Module}
{
    \label{sec:modDesign}
    \begin{figure}[t!]
        \centering
        \includegraphics[width=\linewidth]{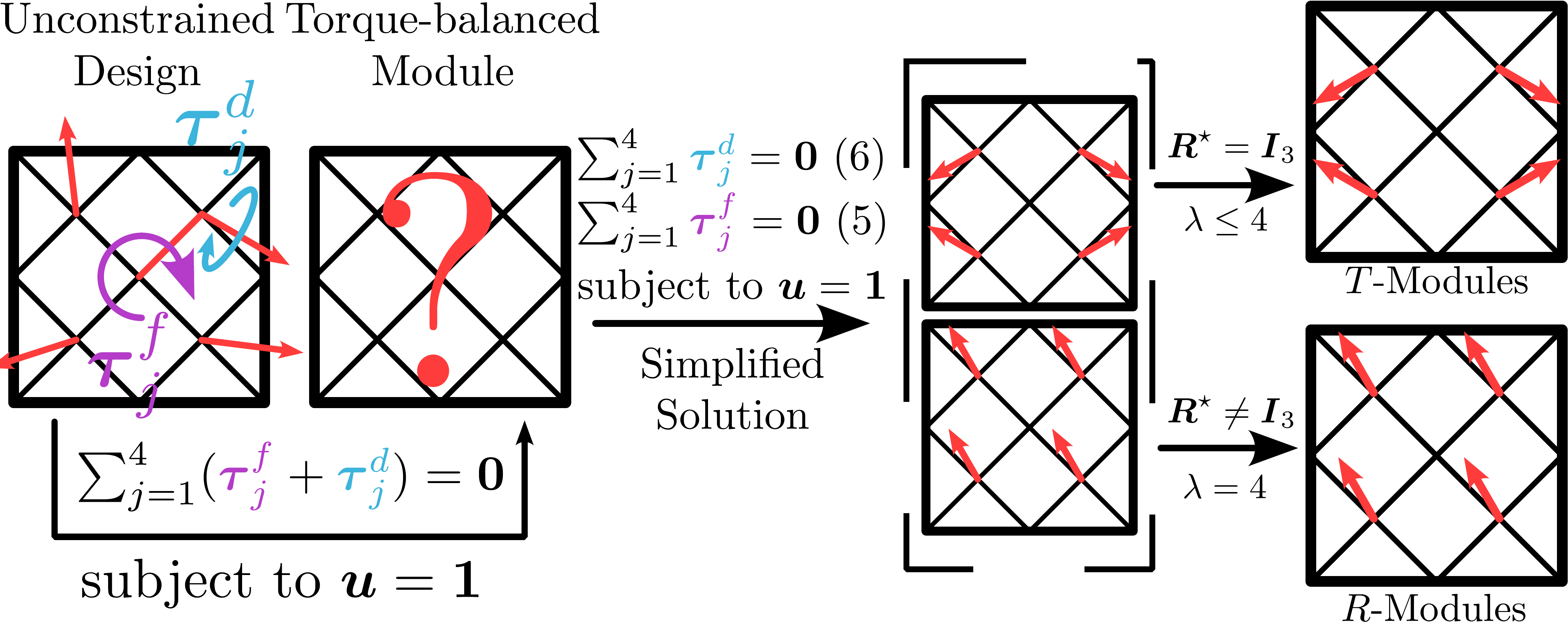}
        \caption{\replytotwo{The procedure of designing torque-balanced modules.}}
        \label{fig:determining-tb-modules}
    \end{figure}
    
    \begin{table}[b!]
        \centering
        \caption{Differences between $R$- and $T$-modules}
        \begin{tabular}{|c|c|c|c|}
        \hline
            \textbf{Module} & \makecell{\textbf{Rotor alignment}\\ \textbf{method}} & \textbf{Parameter} & \textbf{Exp.} \\ \hline
            $R$ & Same direction & $\boldsymbol{R}^\star\in\mathsf{SO(3)}$ & Fig.~\ref{fig:oneUAV}\\ \hline
            $T$ & \makecell{Centrosymmetric} & $\eta\in\mathbb{R}$ & Fig.~\ref{fig:Tmodule}\\ \hline
        \end{tabular}
        \label{tab:RTdiff}
    \end{table}

    A desirable characteristic of a module to fly independently is \emph{torque-balance}, which means that the module can hover without rotating when all rotors generate identical forces.
    \begin{definition}[torque-balanced module]
        A \emph{torque-balanced} module is a module that generates zero torque when all its rotors generate a unit force, \textit{i.e.}, if $\boldsymbol{u}=\boldsymbol{1}$, then $\boldsymbol\tau=\boldsymbol{0}$.
    \end{definition}
    \noindent
    For Problem~\ref{p:moduledesign}, we find the design criterion for a \emph{torque-balanced module}. Given a unit force vector $\boldsymbol{\hat{f}}^\star$, find ${}^{M}\!\boldsymbol{R}_j$ of the four rotors and a scalar value $\lambda>0$, such that when all the rotors generate a unit force, \textit{i.e.}, $\boldsymbol{u}=\boldsymbol{1}$, the force and torque of the module are
    $\boldsymbol{\tau}=\boldsymbol{0}\text{ and } \boldsymbol{f}=\lambda\boldsymbol{\hat{f}}^\star$, where $\boldsymbol{\hat{f}}^\star=\boldsymbol{R}^\star\boldsymbol{\hat z}$ determines the direction of the module force in $\{M_i\}$ specified by the rotation matrix $\boldsymbol{R}^\star$ and the magnitude of the desired force, $\lambda$. We assume $\lambda>0$ because $\lambda=0$ means that the module generates zero force and does not hover.
    $\boldsymbol{R}^\star$ is determined during the module design.}

    Based on~\eqref{eq:ftaustructure}, the local torque of a module $\boldsymbol{\tau}_i$ depends on thrust $\boldsymbol{\tau}^f_{ij}$, and drag  $\boldsymbol{\tau}^d_{ij}$. \replytotwo{Typically, the torque created by the force, $\boldsymbol{\tau}^f_{ij}$, is much greater than by the drag, $\boldsymbol{\tau}^d_{ij}$ because $k_f \gg k_m$. Therefore, a practical solution for $\boldsymbol{\tau}_i=\boldsymbol{0}$ is obtained by solving $\sum\boldsymbol{\tau}^f_{ij}=\boldsymbol{0}$ \emph{and} $\sum\boldsymbol{\tau}^d_{ij}=\boldsymbol{0}$.}
    From \eqref{eq:ftaustructure}, we obtain
    \begin{eqnarray}
        \textstyle\sum_{j=1}^4\boldsymbol{p}_j\times{}^{M}\!\boldsymbol{R}_{j}\boldsymbol{\hat z} &=&\boldsymbol{0}\label{eq:Atf0},\\
        \textstyle\sum_{j=1}^4{}^{M}\!\boldsymbol{R}_{j}(-1)^j\boldsymbol{\hat z}&=&\boldsymbol{0}\label{eq:Atd0},\\
        \textstyle\sum_{j=1}^4{}^{M}\!\boldsymbol{R}_{j}\boldsymbol{\hat z}&=&{\lambda}\boldsymbol{R}^\star\boldsymbol{\hat z},\label{eq:F}
    \label{eq:simplifiedMF}
    \end{eqnarray}
    \noindent
    \replytotwo{where~\eqref{eq:Atf0} ensures $\sum\boldsymbol{\tau}^f_{ij}=\boldsymbol{0}$ and~\eqref{eq:Atd0} ensures $\sum\boldsymbol{\tau}^d_{ij}=\boldsymbol{0}$ under the assumption of unit input force generation $\boldsymbol{u = 1}$.} Note that the discussion is based on one module, meaning that $n = 1$. For conciseness, we omit the module index $i$ in this section and $\boldsymbol{p}_j$ represents the position of the $j$-th rotor in the module frame. We discuss two cases for $\boldsymbol{R}^\star$. 

    \begin{figure}[t!]
        \centering
        \subfloat[{A $T$-module.\label{fig:Tmodule}}
        ]{\centering\includegraphics[width=0.25\linewidth]{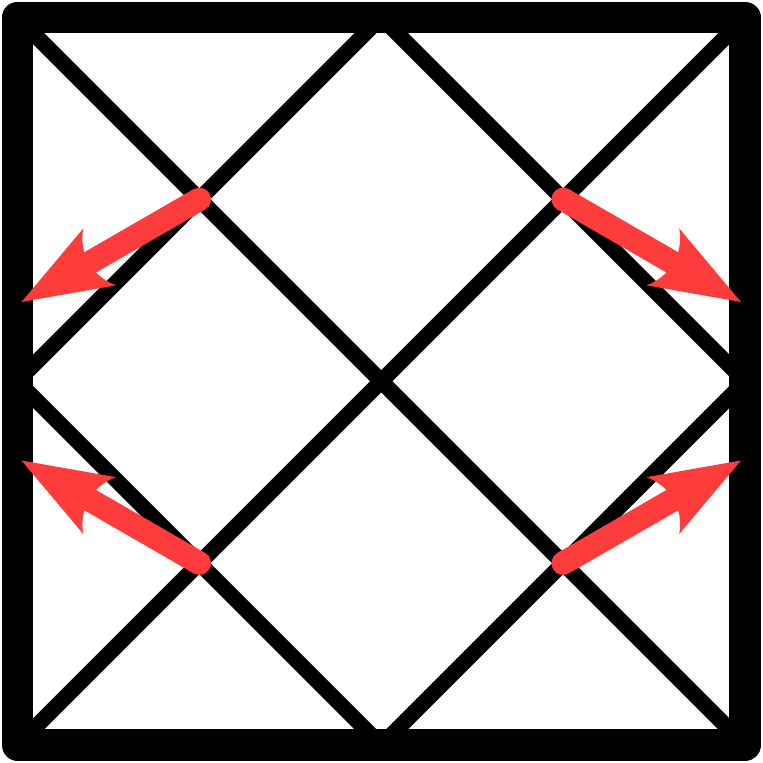}\quad\includegraphics[width=0.45\linewidth]{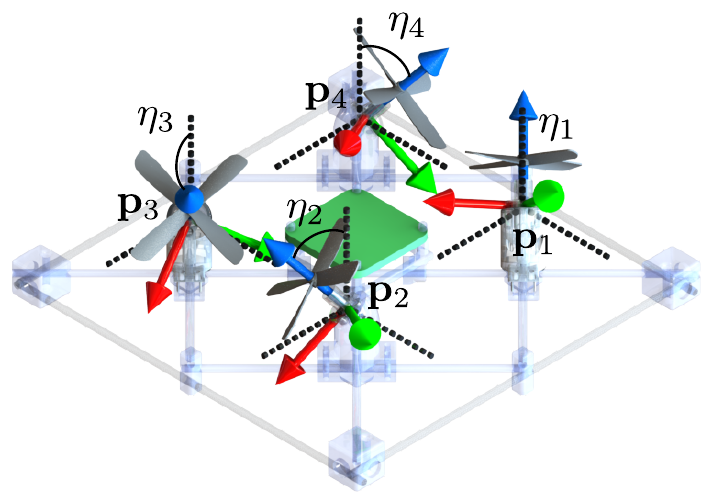}}\\
        \subfloat[{$R$-, $T$-modules. 
        \label{fig:RTModule}}]{\centering\includegraphics[width=0.7\linewidth]{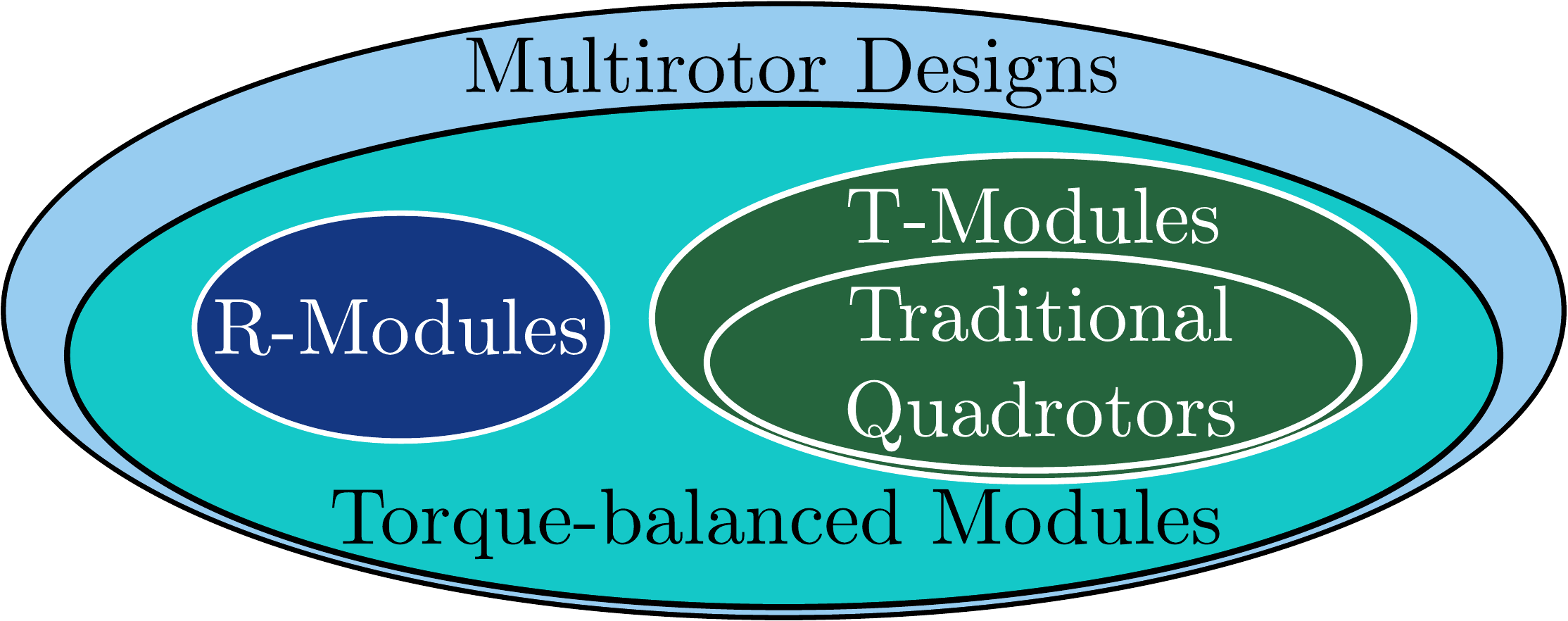}}
        \caption{(a) $T$-modules have their rotors tilted around the arms; (b) A Venn diagram showing the relations between $R$- and $T$-modules.}
    \end{figure}
    
    First, when $\boldsymbol{R}^\star\neq\boldsymbol{I}_3$, the \emph{unique} solution is 
    $^{M}\!\boldsymbol{R}_{j}=\boldsymbol{R}^\star$ for all $j=1,...,4$, ignoring the rotor frames' rotation around their $z$-axis under the constraint of $\boldsymbol{R}^\star\boldsymbol{\hat z}$ lies in the upper hemi-unit sphere.
    This leads us to define a specific type of module that creates a sub-group of torque-balanced modules. 
    
    \begin{definition}[$R$-module]
    An \emph{$R$-module} is a quadrotor module with all its rotors pointing in the same direction in $\{M\}$, specified by the rotation matrix ${}^{M}\!\boldsymbol{R}_j = \boldsymbol{R}^\star, \forall j=1,\dots, 4$.
    \end{definition}
    
    \noindent
    We prove in~\cite{9561016} that an $R$-module is torque-balanced. \replytoall{All rotors of an $R$-module are in the same direction, which infers $\lambda = 4$, ensuring maximum force generation in a single direction.}

    Second, when $\boldsymbol{R}^\star=\boldsymbol{I}_3$, the rotors of a module can point in different directions. Since $R$-modules maximize force in certain directions, we designed a new solution that allows the module to generate a higher torque, which defines the second type of torque-balanced modules.

    \begin{definition}[$T$-Module]
    A \emph{$T$-module} is a module with its rotors tilted at an angle $\eta_j$ around their arm vector $\boldsymbol{p}_j$, $j=1, \dots, 4$. The angles satisfy $\eta_{1} = \eta_{3} = -\eta_{2} = -\eta_{4}$. Making $\eta = \eta_{1}$, we characterize a $T$-module with the parameter $\eta$.
    \end{definition}
    
    \begin{proposition}
    If a module is a $T$-module, defined by the tilt angle $\eta$, the module is a torque-balanced module.
    \label{prop:tbalance}
    \end{proposition}
    \noindent
    We provide the proof of Proposition~\ref{prop:tbalance} in Section 1 of the supplementary file.

    
    Fig.~\ref{fig:oneUAV} and~\ref{fig:Tmodule} show the examples of an $R$- and a $T$-module, respectively. The Venn diagram in Fig.~\ref{fig:RTModule} illustrates the relations between $R$- and $T$-modules, \replytotwo{Fig.~\ref{fig:determining-tb-modules} shows the procedure for designing these two families of torque-balanced modules,} and Table~\ref{tab:RTdiff} summarizes the differences between them. A \emph{module} can refer to a $R$-module or a $T$-module. When a set of modules that include $T$- and/or $R$-modules is assembled into a single rigid structure, it forms an \emph{H-ModQuad structure}, or simply a \emph{structure}. \replytoall{We highlight that a $T$-module with $\eta \neq 0$ indicates ${}^S\!\boldsymbol{R}_{ij} \neq \boldsymbol{I}_3$ in~\eqref{eq:Acomponents}, which results in the cross product having a nonzero component in the $z$-axis. This means that a $T$-module is capable of generating torque in its $z$-axis not only from propeller drag but also from thrust, enhancing the torque generation. Furthermore, we hypothesize that the integration of $T$-modules into a structure allows the structure to achieve higher tilting angles before rotor saturation.}

\section{Actuation Capability Models}{
\label{sec:actuation}
Since the design matrix $\boldsymbol{A}$ contains all the information regarding the configuration of a structure, including the position and orientation of the rotors in ${S}$, our analysis focuses on modeling the actuation capabilities of the structure based on $\boldsymbol{A}$. We draw inspiration from the manipulability ellipsoid~\cite{yoshikawa1985manipulability} and the force envelope~\cite{8206269,Allenspach2020,9476994,hamandi2021design} and develop two models that serve different purposes. The first model, the~\emph{actuation ellipsoid}, provides an efficient analytical insight into the force capabilities of a structure. The second model, the~\emph{actuation polytope}, involves higher computational complexity but can accommodate realistic constraints, which allows designers to evaluate the vehicles against the requirements of aerial tasks.

\subsection{Actuation Ellipsoid}
The work in~\cite{yoshikawa1985manipulability} introduced \emph{manipulability ellipsoids} for robot-arm manipulation by characterizing the response of a manipulator based on its design matrix. 
We extend the concept for H-ModQuad, characterizing the output force in response to rotor input to reveal the direction in which the vehicle is most capable of generating force.
  
As described in~\eqref{eq:Acomponents}, $\boldsymbol{A}$ is composed of two parts, $\boldsymbol{A}_{\boldsymbol{f}}$ and $\boldsymbol{A}_{\boldsymbol\tau}$ which are its first and last three rows, respectively. Note that $\boldsymbol{A_f}$ maps the input rotor force to the total force in $\{S\}$. 
We apply singular value decomposition (SVD) on $\boldsymbol{A_f}$ and use the singular vectors to define the actuation ellipsoid. After normalizing all singular values, we choose the singular vector associated with the largest singular value as the semi-major axis of the actuation ellipsoid, $\boldsymbol{z}_F$, attaching the center of the ellipsoid to the origin of $\{S\}$. The singular vector associated with the second largest singular value is chosen as the second semi-major axis of the actuation ellipsoid, $\boldsymbol{x}_F$, and the singular vector associated with the smallest singular value is chosen as the semi-minor axis of the actuation ellipsoid, $\boldsymbol{y}_F$. 

It is possible that the second and third largest singular values are identical after $\boldsymbol{z}_F$ is found. 
For instance, when $\text{rank}(\boldsymbol{A_f})=1$, all the force vectors from the rotors are co-linear, and the two smallest singular values given by SVD are $0$ and $\boldsymbol{z}_F$ is in the direction of the thrust force. 
In these cases, we set the direction of the second semi-major axis closest to the $x$-axis of $\{S\}$, denoted by $\boldsymbol{\hat x}_S$, i.e., 
$\boldsymbol{x}_F = \Vert\boldsymbol{x}_F\Vert\boldsymbol{\hat x}_F$, where $\Vert\boldsymbol{x}_F\Vert$ is the magnitude of the singular vector associated with the second largest singular value, and
$\boldsymbol{\hat x}_F = \underset{\boldsymbol{v}}{\text{argmax }}\boldsymbol{v\cdot\hat x}_S \text{ such that } \norm{\boldsymbol{v}} = 1, \boldsymbol{v}\cdot\boldsymbol{z}_F=0,$
where $\norm{\boldsymbol{v}} = 1$ emphasizes that the unit vector $\boldsymbol{\hat x}_F$ is the \emph{direction} of the second semi-major axis we choose, and $\boldsymbol{v}\cdot\boldsymbol{z}_F=0$ ensures the major and minor axes of the ellipsoid are orthogonal. The semi-minor axis of the ellipsoid is $\boldsymbol{y}_F = \Vert\boldsymbol{y}_F\Vert\boldsymbol{\hat y}_F$, where $\Vert\boldsymbol{y}_F\Vert$ is the magnitude of the singular vector associated with the smallest singular value, and the direction of the semi-minor axis is $\boldsymbol{\hat y}_F = \boldsymbol{\hat z}_F\times\boldsymbol{\hat x}_F$, $\boldsymbol{\hat z}_F = \frac{\boldsymbol{z}_F}{\Vert\boldsymbol{z}_F\Vert}$. 

The actuation ellipsoid depicts the structure's capability of generating force in different directions, as shown in Fig.~\ref{fig:sideview}. The higher distance between the boundary of the ellipsoid and the origin of $\{S\}$ means that in the direction the structure can generate force more efficiently. \replytoone{We use this model in Section~\ref{sec:control} to find the initial robot frame that the structure can use to maximize hovering efficiency.} 
\label{sec:ellipsoid}
\begin{figure}[t]
    \centering
    \subfloat[{ $\alpha_1 = -\frac{\pi}{6}, \alpha_2 = \frac{\pi}{6}$\label{fig:sideview1}}]{
        \centering
        \includegraphics[clip, trim=-2cm 0cm -6cm 0cm, height=0.32\linewidth]{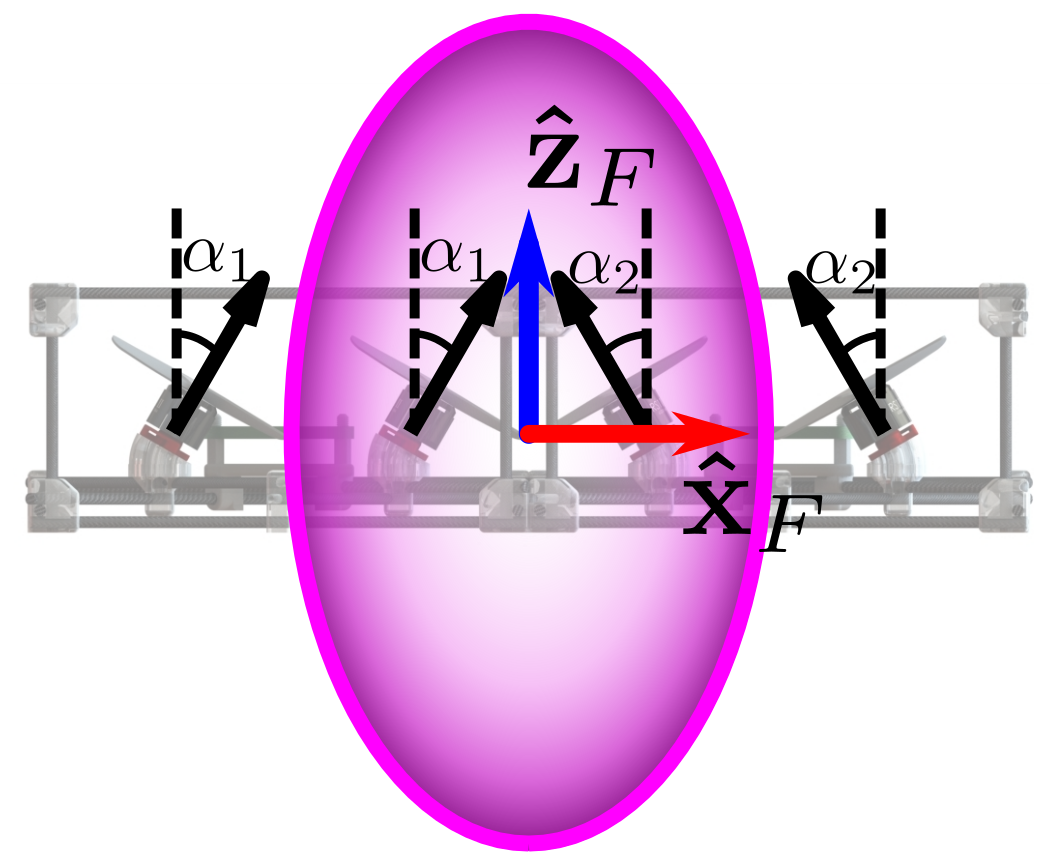}}
    \subfloat[{ $\alpha_1 = \frac{\pi}{4}, \alpha_2 = -\frac{\pi}{4}$\label{fig:sideview2}}]{%
        \centering
        \includegraphics[clip, trim=0cm 0cm -4cm 0cm, height=0.32\linewidth]{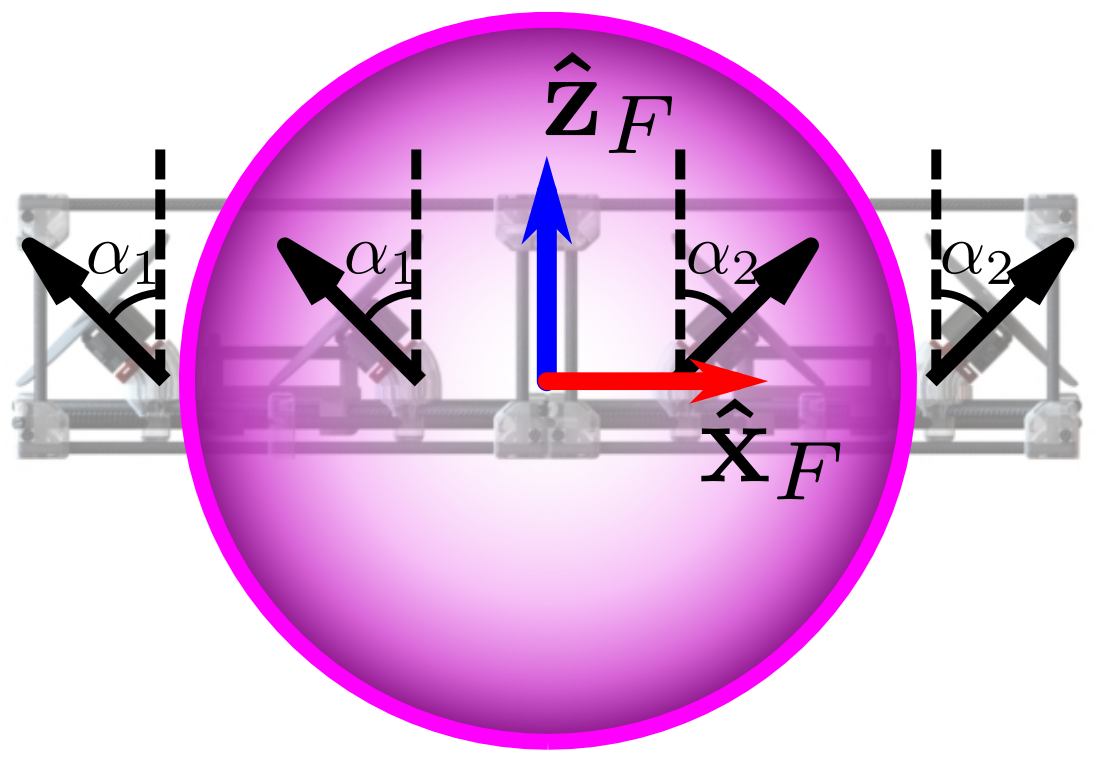}}\\
    \subfloat[{ $\alpha_1 = \frac{\pi}{0}, \alpha_2 = -\frac{\pi}{4}$\label{fig:sideview3}}]{%
        \centering
        \includegraphics[clip, trim=5cm 0cm 5cm -5cm, height=0.35\linewidth]{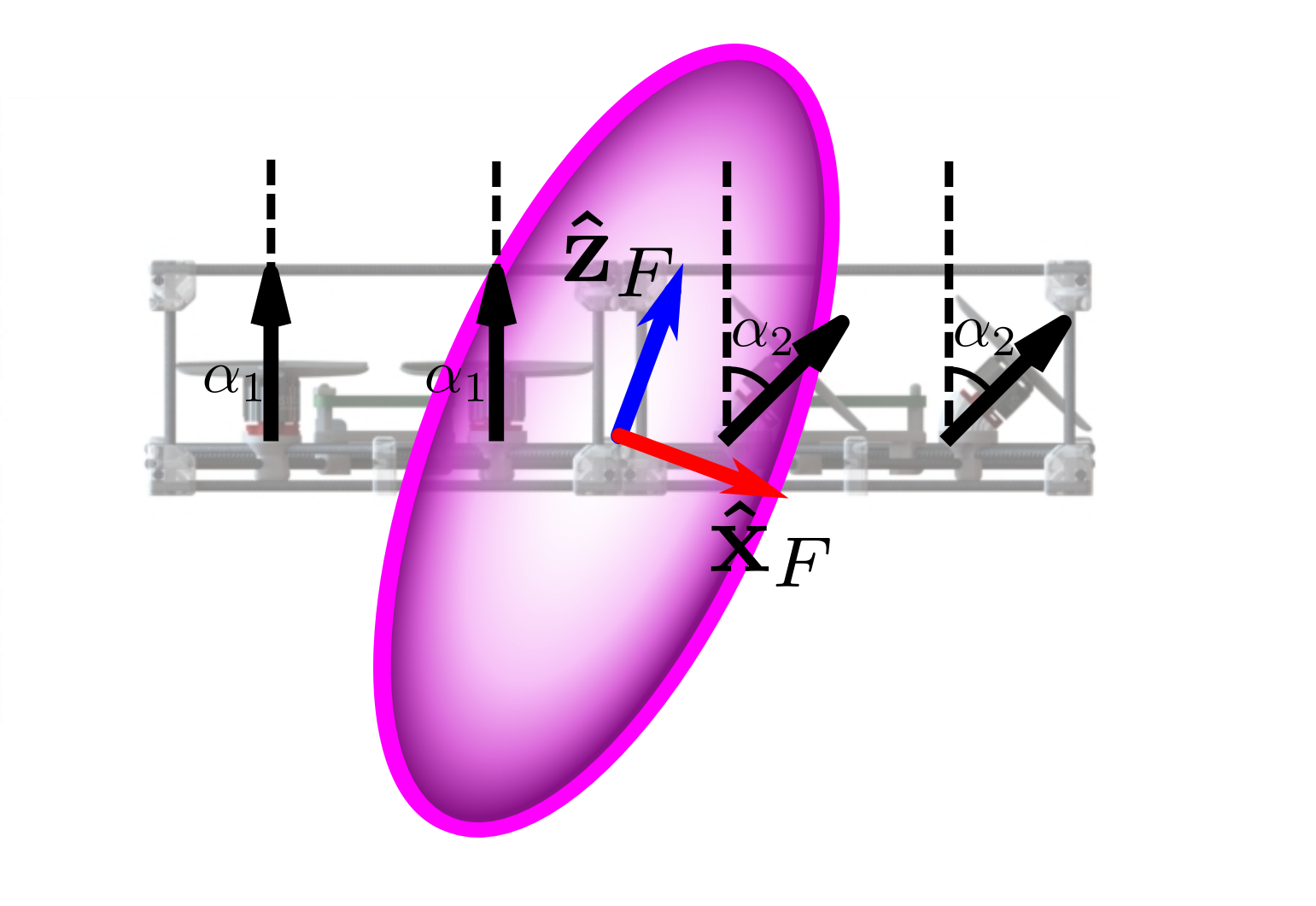}}\qquad
    \subfloat[{ $\alpha_1 = \frac{\pi}{3}, \alpha_2 = -\frac{\pi}{3}$\label{fig:sideview4}}]{%
        \centering
        \includegraphics[clip, trim=0cm -5cm 1cm 0cm, height=0.25\linewidth]{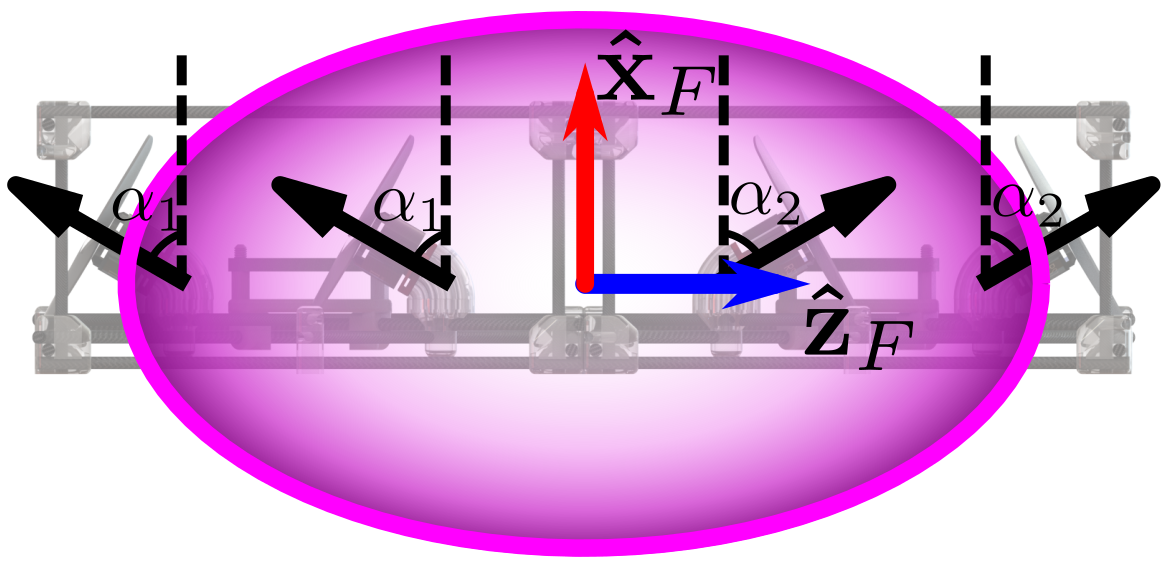}}
    \caption{Side view of four structures, each composed of two types of modules. The black arrows represent the direction of the rotor force. The purple ellipses represent the actuation ellipsoid projected on the $xz$-plane of $\{S\}$. 
    }
    \label{fig:sideview}
\end{figure}

\subsection{Actuation Polytope}
\label{sec:polytope}
The actuation ellipsoid of an H-ModQuad characterizes the capability of the vehicle to generate force in different directions disregarding the constraints on the rotor input, such as the unidirectionality of the rotor rotation and the saturation. These constraints limit the actuation capability of H-ModQuad vehicles to meet aerial task requirements. Since most interaction between an H-ModQuad and objects in the environment involves physical contact, we model the requirement of an aerial task for a vehicle as a set of total wrenches that the vehicle must generate in $\{S\}$.

\begin{definition}[Task Requirement]
    A \emph{task requirement} is a set of wrenches $\mathcal{T} = \{\boldsymbol{w}_{1}, \boldsymbol{w}_{2}, \dots, \boldsymbol{w}_{t}\}\subset\mathbb{R}^6$ that a multirotor vehicle needs to generate to perform the task.
\end{definition}
\replytoone{Each element $\boldsymbol{w}_{t}\in\mathcal{T}$ is a stacked vector of simultaneously required force and torque necessary for completing a task.} Designers must ensure that an aerial vehicle is capable of performing its intended task. Typical evaluation of actuation capability involves constructing and testing of prototypes, which can be time-consuming despite the modularity of H-ModQuad. Therefore, we develop a method to understand the actuation capability of a structure before assembly and experimental evaluation. 
We develop \emph{actuation polytope} as the abstraction of the actuation capability for multirotor vehicles composed of unidirectional rotors with fixed poses in a vehicle. Inspecting the actuation polytope of a vehicle, we understand its actuation capabilities and can further reconfigure the modules to achieve better compatibility with a task.

Based on \eqref{eq:MAu} and \eqref{eq:Acomponents}, the design matrix $\boldsymbol{A}$ converts the rotor force to the total wrench in $\{S\}$. Since the unidirectional rotors can generate a maximum thrust force $f_{max}$, the set of all possible wrenches that a structure can generate is
\begin{equation}
    \mathcal{W} = \left\{\boldsymbol{A}\boldsymbol{u}\:\vert\: 0\preceq \boldsymbol{u}\preceq f_{max}, \boldsymbol{u}\in\mathbb{R}^{4n}\right\},
    \label{eq:polytopeSet}
\end{equation}
where ``$\preceq$'' stands for element-wise comparison and the lower bound $0$ emphasizes the motor unidirectionality. The constraints on the input $0\preceq\boldsymbol{u}\preceq f_{max}$ represent the intersection of $2\cdot4n$ half-spaces in the $\mathbb{R}^{4n}$ input space, which is a convex polyhedron~\cite{boyd2004convex}. Since the polyhedron is only constrained with upper and lower bounds along each axis, it is a hypercube. The linear mapping of the hypercube with the matrix $\boldsymbol{A}$ to the set of all possible wrenches $\mathcal{W}\subset\mathbb{R}^6$ preserve the convexity and boundedness. Thus, $\mathcal{W}$ is a convex polytope in $\mathbb{R}^6$ wrench space, which we define as \emph{actuation polytope}. 

The actuation polytope in~\eqref{eq:polytopeSet} is an infinite set in $\mathbb{R}^6$. Since the task requirement $\mathcal{T}$ is defined as a set of required wrenches, we can compare each element of $\mathcal{T}$ with $\mathcal{W}$ of a structure. Given a required wrench $\boldsymbol{w}_r\in\mathcal{T} = \chi_r\boldsymbol{\hat w}$, where $\chi_r$ and $\boldsymbol{\hat w}$ are the magnitude and unit vector of $\boldsymbol{w}_r$, we find the maximum magnitude, $\chi$, of the wrench the vehicle can generate in the direction of $\boldsymbol{w}_r$ by solving the optimization problem
\begin{equation}
    \begin{aligned}
    & \underset{\boldsymbol{u}}{\text{max}}
    & & \chi,\\
    & \text{subject to}
    & & 0\preceq\boldsymbol{u}\preceq f_{max},\\
    & & & \boldsymbol{Au} = \chi\boldsymbol{\hat w}.
    \end{aligned}
    \label{eq:convexopt}
\end{equation}
Since $\boldsymbol{0}\in\mathcal{W}$ always holds true by setting $\boldsymbol{u=0}$ and $\mathcal{W}$ is convex, if $\chi\geq\chi_r$, then $\boldsymbol{w}_r\in\mathcal{W}$, meaning that the vehicle is able to satisfy the task requirement $\boldsymbol{w}_r$. 

The boundary of $\mathcal{W}$ provides an intuitive visualization of the actuation capabilities due to its convexity. By solving \eqref{eq:convexopt} for all $\boldsymbol{\hat w}$ that is on a unit 6-sphere, \textit{i.e.,} finding the greatest wrenches in $\mathcal{W}$ in all possible directions, we obtain the boundary of $\mathcal{W}$. By projecting the boundary into the $\mathbb{R}^3$ force space, we obtain the \emph{force envelope} of a structure. Similarly, we can obtain the \emph{torque envelope} by projecting the boundary of $\mathcal{W}$ into the torque space. However, in reality, the force and torque that a structure generates are coupled. For example, when hovering, a structure needs to maintain zero torque while keeping the negative gravity in its force envelope, which limits the possible forces it can generate, rendering achievable only a fraction of the force envelope. 
We highlight that such a coupling relationship can be taken into account by incorporating additional constraints in Eq.~\eqref{eq:convexopt}. 

Taking the example of determining the maximum tilt angle of a structure without rotation, we add a constraint $\boldsymbol{A_\tau u = 0}$ to~\eqref{eq:convexopt} and project the polytope into the $\mathbb{R}^3$ force space. This new constrained polytope of the force envelope, namely, force polytope, reveals the available forces the structure can generate without tilting. Since it is attached to $\{S\}$, the force vector that compensates gravity $\boldsymbol{g}$ must rotate inside the force polytope depending on the structure's attitude. To verify whether the structure is capable of achieving the desired attitude ${}^W\!\!\boldsymbol{R}^d_S$ while hovering, we calculate the force required to generate in~$\{S\}$, $\boldsymbol{f}^d = ({}^W\!\!\boldsymbol{R}^d_S)^\top\boldsymbol{g}$. This force being the only element in the task requirement, we compare its magnitude~$\Vert\boldsymbol{f}^d\Vert$ with the maximum force that the structure can generate in the direction of $\boldsymbol{f}^d$ by replacing $\boldsymbol{\hat w}$ with $\frac{\boldsymbol{f}^d}{\Vert\boldsymbol{f}^d\Vert}$, and $\boldsymbol{A}$ with $\boldsymbol{A_f}$ in~\eqref{eq:convexopt}. If the maximum force magnitude $\chi > \Vert\boldsymbol{f}^d\Vert$, then we confirm the structure is able to hover at a desired attitude ${}^W\!\!\boldsymbol{R}^d_S$.
\replytoall{Using this method, we calculate that the maximum tilt angle of the structure used in Experiments 4, 5, and 6 is $37.9^\circ$ in both roll and pitch, compared to $12^\circ$ for the structure used in \replytotworoundtwo{Experiment 3}. These theoretical results are tested in the experiments and compared with our analysis.}

When designing a structure, we examine its force polytope to determine whether the structure satisfies the intended task requirements. 
In~\cite{10160555}, we show how to leverage the actuation polytope to find the optimal configuration of a growing structure given a limited number of homogeneous $T$-modules.
Table 1 in the supplementary file shows some different structure configurations suitable for different tasks. 
\begin{figure*}[t!]
    \centering
    {\includegraphics[width=0.9\linewidth]{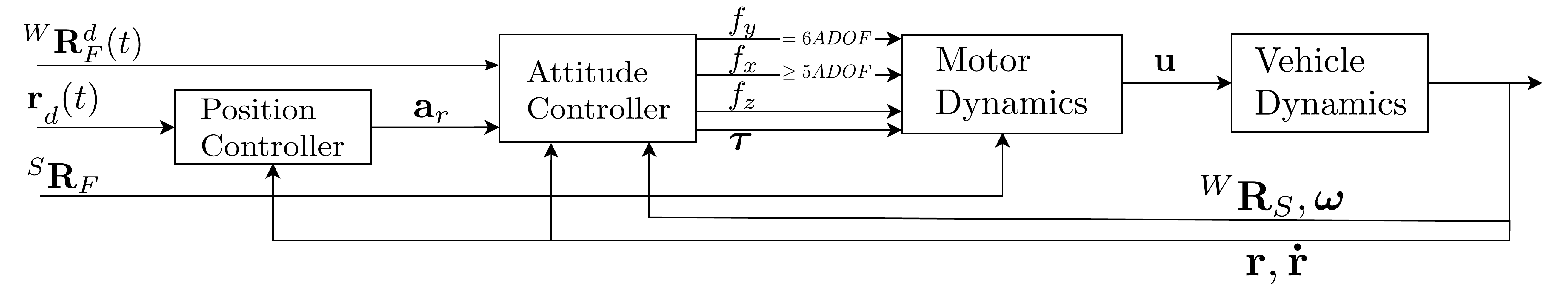}
    \caption{Control Diagram: We track the desired pose of $\{F\}$, $\boldsymbol{r}^d(t), {}^W\!\!\boldsymbol{R}^d_F(t)$ using feedback on the pose of the structure frame $\{S\}$, ${}^W\!\!\boldsymbol{R}_S(t), \boldsymbol{r}(t)$. The Motor Dynamics block is characterized by the matrix $(\boldsymbol{DA})^\dagger$. The rotation matrix from $\{S\}$ to $\{F\}$, ${}^{S}\!\boldsymbol{R}_F$, is computed based on the structure configuration.}
    \label{fig:diagram}}
\end{figure*}

\section{Control}{
\label{sec:control}

Our approach for trajectory tracking control is composed of two parts: position and attitude. {Fig.~\ref{fig:diagram} shows an overview of the control strategy in a centralized way where all modules in the structure are controlled based on a trajectory function and a single stream of measurements. } 

Given a desired trajectory function of time, $\boldsymbol{r}^d$, a classical PID trajectory tracking control can compute an acceleration vector $\boldsymbol{a_r}$ by deriving the tracking errors, $\boldsymbol{e_r} = \boldsymbol{r}^d-\boldsymbol{r}$, and, $\boldsymbol{e_{v}} = \boldsymbol{\dot r}^d-\boldsymbol{\dot r}\label{eq:ev}$, using the feed-forward equation,
\begin{equation}
        \boldsymbol{a_r} = \boldsymbol{K_re_r+K_ve_v}+\replytotwo{\boldsymbol{K_{i,r}}\int \boldsymbol{e_r}dt}+g\boldsymbol{\hat z}+\boldsymbol{\ddot r}^d\label{eq:fd},
\end{equation}
where $\boldsymbol{K_r}$, $\boldsymbol{K_v}$, and $\boldsymbol{K_{i,r}}$ are positive-definite gain matrices.

In order to efficiently achieve the desired acceleration based on the structure configuration, we introduce the concept of $F$-frame using the actuation ellipsoid. We note that in the direction of the semi-major axis of the ellipsoid, $\boldsymbol{\hat z}_F$, the structure is able to maximize force generation, and in the direction of the second semi-major axis, $\boldsymbol{\hat x}_F$, the structure maximizes its force generation in the normal plane of $\boldsymbol{\hat z}_F$.

\begin{definition}[$F$-frame]
    The $F$-frame of a structure, denoted by $\{F\}$, 
    is a coordinate frame with its origin at the origin of $\{S\}$;
    the $z$-axis, $\boldsymbol{\hat z}_F$, points towards the direction where the structure can generate its maximum thrust;
    the $x$-axis, $\boldsymbol{\hat x}_F$, points towards the direction where the structure can generate its maximum thrust on the normal plane of $\boldsymbol{\hat z}_F$.
\end{definition}
Therefore, we obtain the rotation matrix from $\{S\}$ to the $F$-frame, ${}^{S}\!\boldsymbol{R}_F=\left[\boldsymbol{\hat x}_F\; \boldsymbol{\hat y}_F\; \boldsymbol{\hat z}_F\right],$ where $\boldsymbol{\hat y}_F = \boldsymbol{\hat z}_F\times\boldsymbol{\hat x}_F$. To maximize the hover efficiency, we derive the attitude tracking errors based on the $F$-frame instead of $\{S\}$.

We compute the attitude error in $\mathsf{SO(3)}$ with respect to the $F$-frame based on~\cite{5717652}. Since the structure has attitude ${}^W\!\!\boldsymbol{R}_F$ and angular velocity $\boldsymbol{\omega}$, the angular tracking error is
{\footnotesize
\begin{eqnarray}
        \boldsymbol{e_R} &=& \frac{1}{2}\left(\left({}^W\!\!\boldsymbol{R}_F^d\right)^{\top}{}^W\!\!\boldsymbol{R}_S{}^{S}\!\boldsymbol{R}_F-\left({}^W\!\!\boldsymbol{R}_S{}^{S}\!\boldsymbol{R}_F\right)^{\top}{}^{W}\!\boldsymbol{R}_F^d\right)^{\vee}\nonumber,\\
        \boldsymbol{e}_{\boldsymbol\omega} &=& \boldsymbol\omega - \left({}^W\!\!\boldsymbol{R}_S{}^{S}\!\boldsymbol{R}_F\right)^{\top}{}^W\!\!\boldsymbol{R}_F^d\boldsymbol\omega^d, \label{eq:eattitude}
    \label{eq:attitudeerror}
\end{eqnarray}
}
where the ``\textit{vee}'' operator, $\vee$, maps a skew symmetric matrix to $\mathbb{R}^3$. 
In order to maximize the actuation efficiency of the structure, our controllers drive the attitude of $\{F\}$ to the desired attitude
${}^W\!\!\boldsymbol{R}_F^d$.
As a result, the attitude of $\{S\}$ converges to ${}^W\!\!\boldsymbol{R}_F^d{}^{S}\!\boldsymbol{R}_F^\top$.
Then, the necessary angular acceleration to compensate the attitude error is
\begin{eqnarray}
    \boldsymbol{a_R} = -\boldsymbol{K_Re_R}-\boldsymbol{K_{\boldsymbol\omega} e_{\boldsymbol\omega}}\label{eq:tau},
\end{eqnarray}
where $\boldsymbol{K_R}$ and $\boldsymbol{K}_{\boldsymbol\omega}$ are positive-definite gain matrices.
Different from the approach proposed by~\cite{5717652}, our attitude error is not with respect to the structure frame, $\{S\}$, but the $F$-frame. Our control policy depends on $\text{rank}(\boldsymbol{A})$. We study each case.

\subsubsection{4 ADoF}
For a structure with rank$(\boldsymbol{A})=4$, as shown in Fig.~\ref{fig:HModQuads}(a), we control the position and yaw angle of the structure using a geometric controller~\cite{5717652}.
We project the desired thrust vector on the $z$-axis of $\{F\}$ to increase hover efficiency.
Given a desired acceleration $\boldsymbol{a_r}$ and a desired yaw $\psi^d$, the desired attitude is ${}^W\!\!\boldsymbol{R}_F^d = [\boldsymbol{\hat x}^d\;\boldsymbol{\hat y}^d\;\boldsymbol{\hat z}^d]$, where
\begin{alignat}{3}
    \boldsymbol{\hat z}^d &= \frac{\boldsymbol{a_r}}{\Vert\boldsymbol{a_r}\Vert}
    , & & \qquad&\boldsymbol{\hat x}^c &= [\cos(\psi^d), \sin(\psi^d), 0]^{\top}%
    \nonumber
    \\
    \boldsymbol{\hat y}^d &= \frac{\boldsymbol{\hat z}^d\times\boldsymbol{\hat x}^c}{\Vert\boldsymbol{\hat z}^d\times\boldsymbol{\hat x}^c\Vert},
    & & \qquad&
    \boldsymbol{\hat x}^d &= \boldsymbol{\hat y}^d\times\boldsymbol{\hat z}^d. 
    \label{eq:direction4}
\end{alignat}

\subsubsection{5 ADoF}
A structure with $n\geq2$ and rank($\boldsymbol{A}$)=$5$ has 5 ADoF, as shown in Fig.~\ref{fig:HModQuads}(b). 
We choose to use the additional ADoF in $\boldsymbol{\hat x}_F$ to track the pitch of $\{F\}$, $\theta^{d}$. The desired attitude is $^W\!\!\boldsymbol{R}^d_F=[\boldsymbol{\hat x}^d\,\:\boldsymbol{\hat y}^d\,\:\boldsymbol{\hat z}^d]$ where 
{\small
\begin{alignat}{3}
        \boldsymbol{\hat z}^c &= \frac{\boldsymbol{a_r}}{\Vert\boldsymbol{a_r}\Vert}
        , & & \qquad&\boldsymbol{\hat x}^d &= \text{Rot}(z, \psi^d)\:\text{Rot}(y, \theta^d)\: \boldsymbol{\hat x},
        \nonumber\\
        \boldsymbol{\hat y}^d &= \frac{\boldsymbol{\hat z}^c\times\boldsymbol{\hat x}^d}{\Vert\boldsymbol{\hat z}^c\times\boldsymbol{\hat x}^d\Vert},
        & & \qquad& \boldsymbol{\hat z}^d&=\boldsymbol{\hat x}^d\times\boldsymbol{\hat y}^d,
    \label{eq:direction5}
\end{alignat}}
where the operators $\text{Rot}\left(x, \cdot\right), \text{Rot}\left(y, \cdot\right), \text{ and }\text{Rot}\left(z, \cdot\right)$ convert the Euler angles in roll, pitch, and yaw into rotation matrices, respectively. Different from~\eqref{eq:direction4}, we obtain $\boldsymbol{\hat x}^d$ in~\eqref{eq:direction5} by applying the desired yaw and pitch on $\boldsymbol{\hat x}$. $\boldsymbol{\hat z}^c$ is projected in the normal direction of $\boldsymbol{\hat x}^d\boldsymbol{\hat y}^d$-plane to acquire the desired force in the $xz$-plane of $\{F\}$, in which all possible forces that the structure can generate are co-planar. This method ensures that the desired attitude captures the desired pitch angle.

\begin{figure}[t]
    \centering
    \subfloat[{ 4 ADoF\label{fig:4DoFHModQuad}}]{%
       \includegraphics[width=0.40\linewidth, 
        trim={9cm 2cm 3cm 3cm},clip]{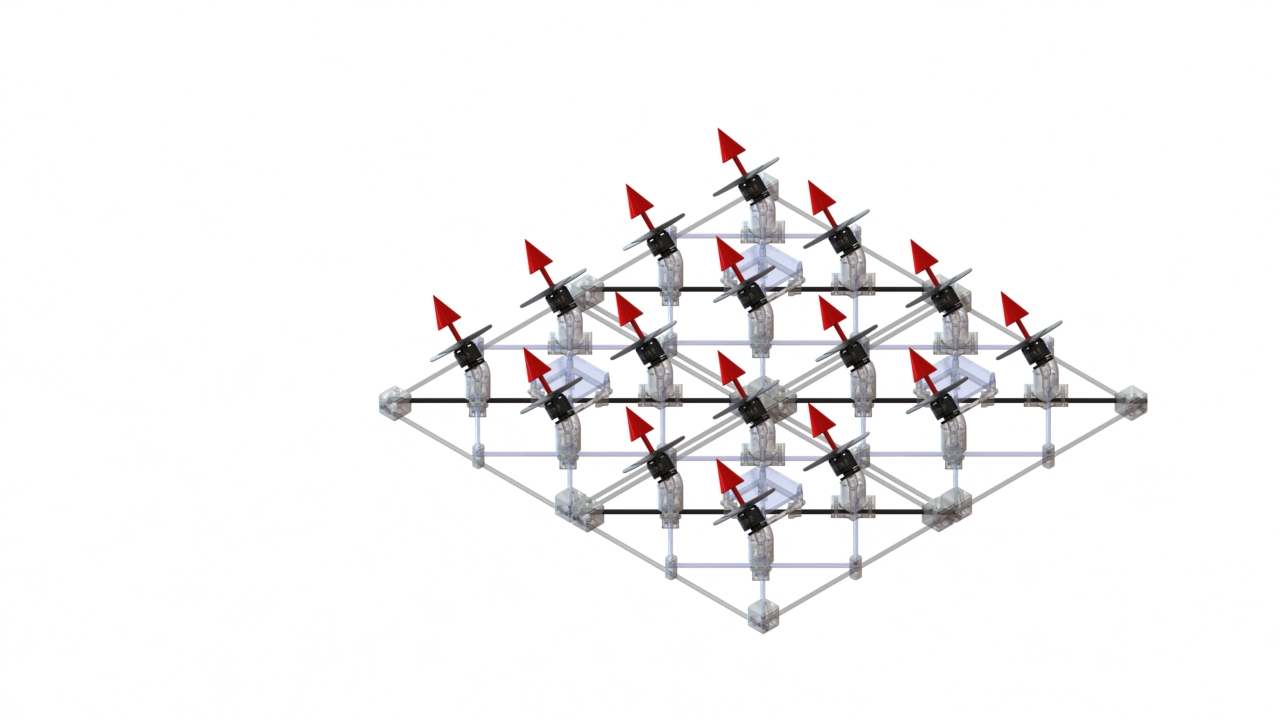}}
    \quad
    \subfloat[{ 5 ADoF\label{fig:5DoFHModQuad}}]{%
        \includegraphics[width=0.40\linewidth, 
        trim={9cm 2cm 3cm 3cm},clip]{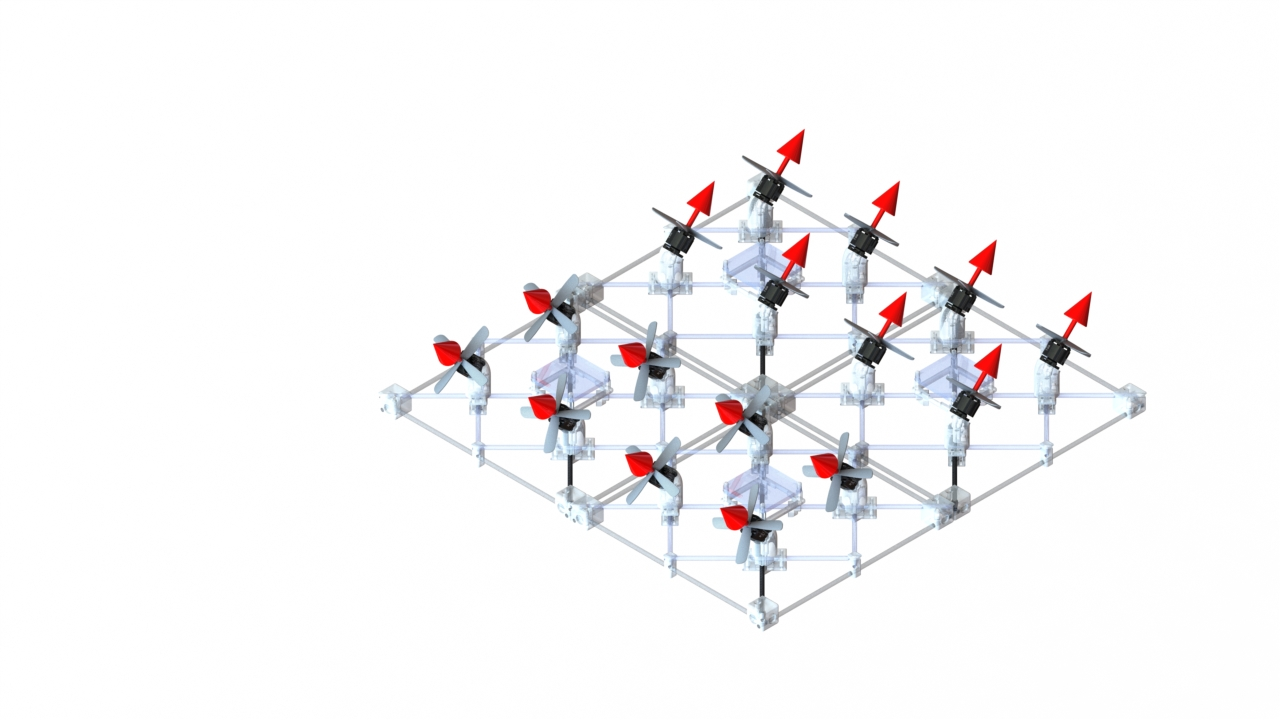}}\\
    
    \subfloat[{ Full actuation\label{fig:6DoFHModQuad}}]{%
        \includegraphics[width=0.40\linewidth, 
        trim={0.5cm 0.cm 0cm 1.2cm},clip]{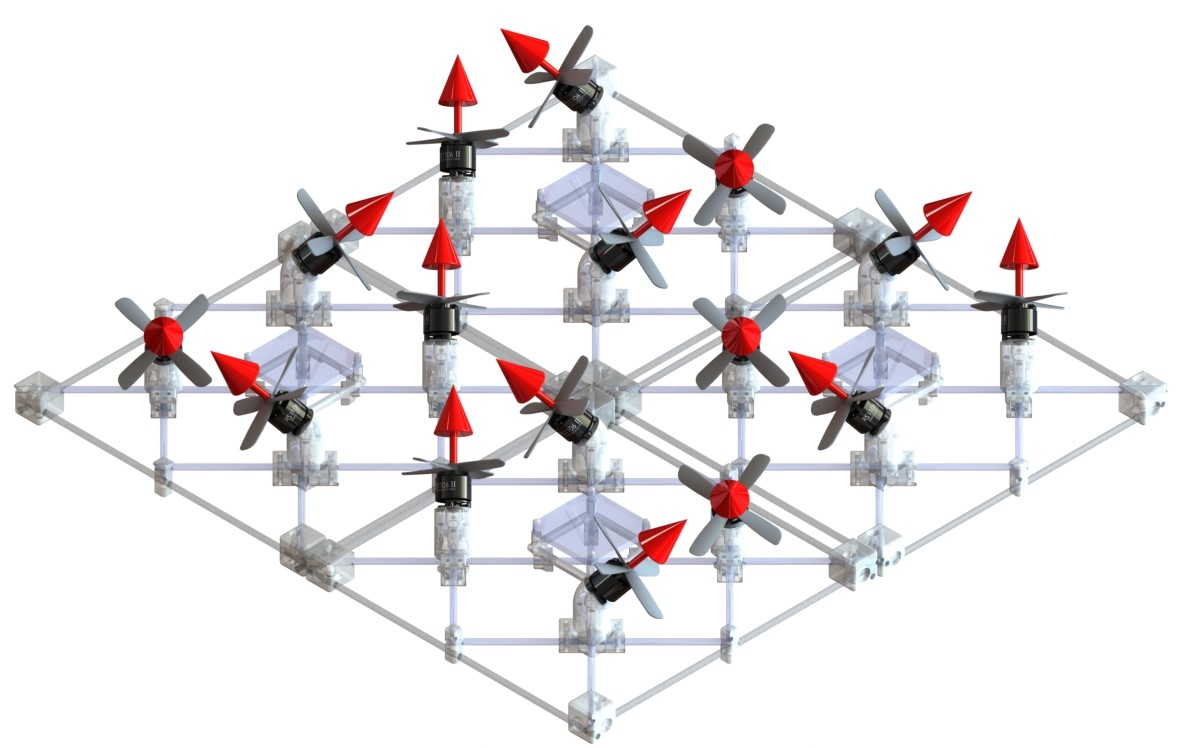}}
    \quad
    \subfloat[{ Omnidirection\label{fig:omniHModQuad}}]{%
        \includegraphics[width=0.36\linewidth, 
        trim={-2cm 0.cm 1cm 1.0cm},clip]{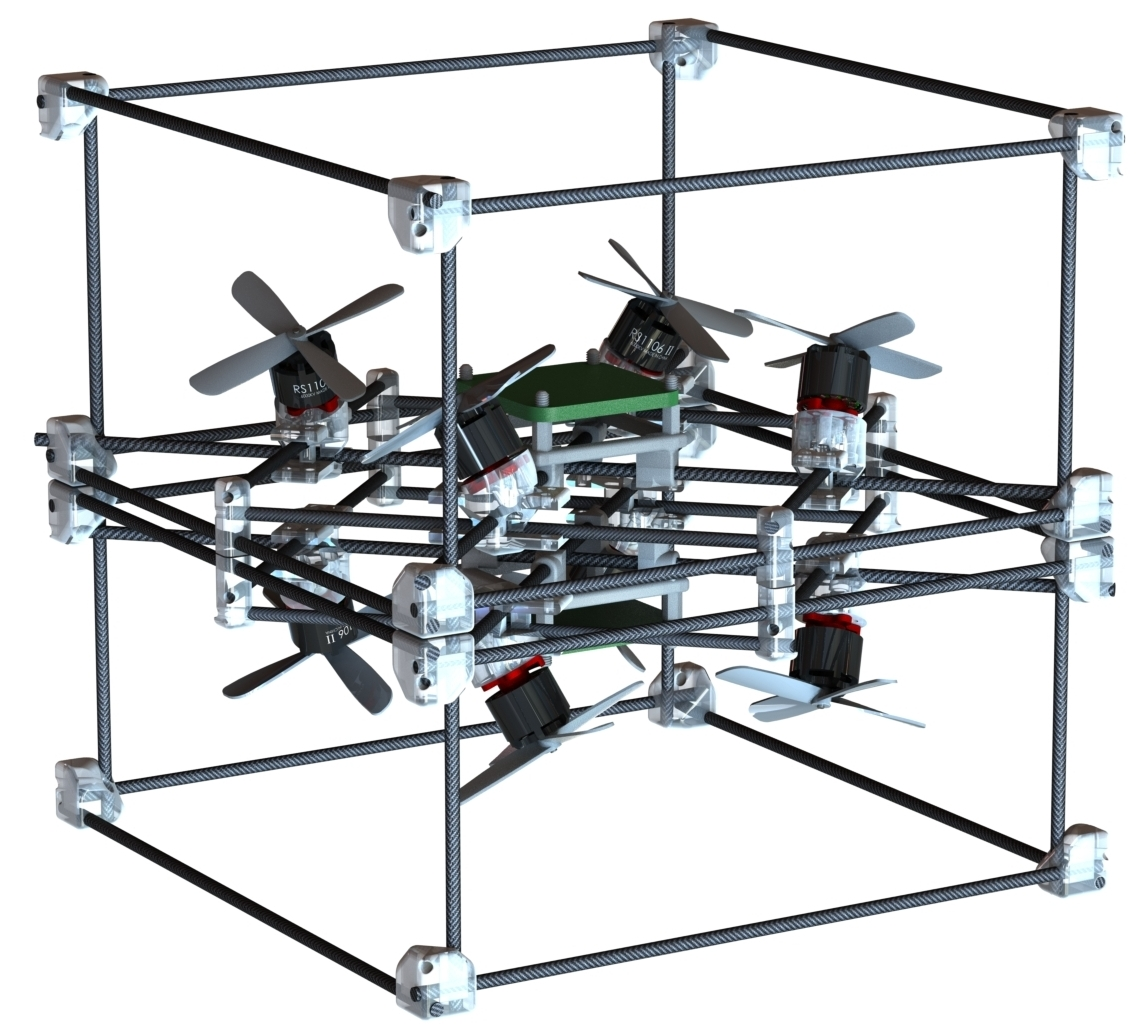}}
    \caption{Four H-ModQuad configurations. In (a), the vehicle composed of four identical $R$-modules has 4 ADoF. In (b), the vehicle is composed of two pairs of $R$-modules, each pair having the same $\boldsymbol{R}^\star$, thus having 5 ADoF. In (c), the vehicle is fully-actuated, composed of four $T$-modules. In (d), the vehicle composed of two $T$-modules back to back is omnidirectional based on the definition given by~\cite{8281444}.}
    \label{fig:HModQuads}
\end{figure}

\subsubsection{6 ADoF}
For a structure with rank$(\boldsymbol{A}) = 6$, as shown in Fig.~\ref{fig:HModQuads}(c) and (d), the vehicle is fully actuated. Therefore, the desired attitude $^W\!\!\boldsymbol{R}^d_F$ is chosen independently.

Substituting $^W\!\!\boldsymbol{R}^d_F$ in~\eqref{eq:eattitude} and replacing the errors in~\eqref{eq:tau}, we obtain the desired angular acceleration $\boldsymbol{a_R}$. Combining $\boldsymbol{a_R}$ with the desired positional acceleration vector~$\boldsymbol{a_r}$, the desired acceleration vectors become $\boldsymbol{a}=\left[\boldsymbol{a_r}^\top, \boldsymbol{a_R}^\top\right]^\top$. To implement feedback linearization, we let the structure to generate wrench $\boldsymbol{w}$ in $\{F\}$ based on~\eqref{eq:Lagrange},
\begin{equation}
    \boldsymbol{w} = \begin{bmatrix}
        {}^W\!\!\boldsymbol{R}_F^\top & \boldsymbol{0}\\
        \boldsymbol{0} & \boldsymbol{J}
    \end{bmatrix}
    \left(
    \boldsymbol{a}
    + \begin{bmatrix}
        \boldsymbol{0}\\
        \boldsymbol\omega\times \boldsymbol{J}\boldsymbol\omega
    \end{bmatrix}\right).
    \label{eq:wrench}
\end{equation}

To generalize to all cases of rank$(\boldsymbol{A})$, we modify \eqref{eq:MAu} into
    $\boldsymbol{Dw} = \boldsymbol{DAu},$
with the \emph{Dimensioning matrix}
\begin{eqnarray}
    \boldsymbol{D} = \left\{\begin{matrix}
    \begin{bmatrix}
        \boldsymbol{0}_{4\times2} & \boldsymbol{I}_4
    \end{bmatrix}, & \text{rank(}\boldsymbol{A}\text{)} = 4,\\
    \ & \ \\
    \begin{bmatrix}
        \begin{bmatrix}
            1 & 0
        \end{bmatrix} & \boldsymbol{0}_{1\times4} \\
        \boldsymbol{0}_{4\times2} & \boldsymbol{I}_4
    \end{bmatrix}, & \text{rank(}\boldsymbol{A}\text{)} = 5,\\
    \ & \ \\
    \boldsymbol{I}_6, & \text{rank(}\boldsymbol{A}\text{)} = 6,
    \end{matrix} \right.
\end{eqnarray}
where $\boldsymbol{I}_a\in\mathbb{R}^{a\times a}$ stands for an $a\times a$ identity matrix, and $\boldsymbol{0}_{a\times b}\in\mathbb{R}^{a\times b}$ represents a zero matrix of size $a\times b$.
We can then calculate the desired input vector by applying module-wise Moore-Penrose inverse on matrix $\boldsymbol{DA}$, 
\begin{equation}
    \boldsymbol{u}_i = (\boldsymbol{DA})^\dagger_i\boldsymbol{Dw},\label{eq:distributedMD}
\end{equation}
where $i=1, \dots, n$, and $(\boldsymbol{DA})^\dagger_i$ is the submatrix of $(\boldsymbol{DA})^\dagger$ composed of its $\left(4i-3\right)$-th to $4i$-th rows, associated with the $i$-th module. The ``Motor Dynamics'' in Fig.~\ref{fig:diagram} refers to this operation.
\replytotwo{The controller is exponentially stable and its proof follows the same logic as in~\cite{5717652} by replacing $^W\!\!\boldsymbol{R}_S$ with $^W\!\!\boldsymbol{R}_F$.}
}

\section{Evaluation}{
\label{sec:eva}
We use real-robot experiments to evaluate the H-ModQuad design and validate control strategies by measuring the tracking error to the desired trajectories. We design six different experiments, where a structure follows a specific trajectory.~\footnote{Narrated experiment recordings and additional simulations with up to 16 modules in the structure showing the scalability of the system can be found at~\url{https://tinyurl.com/hmodquad-evaluation}. The recorded rosbag files are accessible at~\url{https://tinyurl.com/H-ModQuad-data}.}

\subsection{H-ModQuad robots} {
\label{sec:realdesign}
\subsubsection{Design}
\label{sec:subdesign}

We build the H-ModQuad modules based on Crazyflie quadrotors. Each module has four brushless motors and a Crazyflie Bolt control board,
weighing $135$~g including a 2-cell LiPo battery, with a payload capability of $128$ g.\footnote{A complete list of components and assembly instruction can be found at \url{https://docs-quad.readthedocs.io/en/latest/assemble/design.html}; 
A picture of the $T$-module prototype can be found at Fig. 1 of the supplementary file.}
\replytothree{The docking mechanism consists of permanent magnets mounted at the 3D-printed module frame corners. The disc-shaped light-weight Neodymium magnets create rigid connection between modules. Our experiments assume the docking is completed and focus on the behaviors of the assembled structures.}
}

\subsubsection{Localization and Communication}
\label{sec:subcommunication}
In the experimental testbed, we use the Crazyflie-ROS framework~\cite{CrazyflieROS} to command the robots. We modify the Crazyflie firmware.~\footnote{The modified Crazyflie-ROS framework source code is available on GitHub at~\url{https://github.com/swarmslab/customized_Crazyflie_ros} 
and the modified firmware at~\url{https://github.com/swarmslab/modquad-firmware/tree/HModQuad} 
for the $2\times2$ structure in experiments 4, 5, and 6.} 
For localizing the quadrotors, we use the Optitrack system operating at 120 Hz. The IMU onboard measures the angular velocities and linear accelerations. The IMU and tracking system readings are combined with an extended Kalman filter onboard~\cite{doi:10.2514/1.G000848}. 
In the experiments, the structure composes a rigid body in the Optitrack system, of which the pose is transmitted to all the modules. 
The central station broadcasts commands in the form of desired position, orientation, linear and angular velocity of the structure. All modules run the position and attitude controller independently based on their own state estimation and 
calculate the input as given in~\eqref{eq:distributedMD} based on the distributed motor dynamics $(\boldsymbol{DA})^\dagger_i\boldsymbol{D}$ that we implement prior to the experiments. 
Therefore, there is no necessity for inter-robot communication.

\begin{table}[b]
    \centering
    \label{tab:exptable}
    \caption{Module types, ADoF and trajectory types of the experiments.}
    \begin{tabular}{|c|c|c|c|c|}
    \hline
        Exp. & $R$-modules & $T$-modules & \#ADoF & Trajectory  \\ \hline
        1 & 1 & 0 & 4 & Helix  \\
        2 & 2 & 0 & 5 & Rectangle \\
        3 & 4 & 0 & 6 & Rectangle\\
        4 & 0 & 4 & 6 & $\sin(t)$ for $\phi$\\
        5 & 0 & 4 & 6 & Input $\phi$\\
        6 & 0 & 4 & 6 & Rectangle\\ \hline
    \end{tabular}
\end{table}
\subsection{Experiments}{
In Experiments 1, 2, and 3, we validate our method on different structures composed of $R$-modules. In Experiments 4, 5, and 6, we show the performance of a fully actuated structure composed of $T$-modules.
\label{sec:exp}
}

\subsubsection{
Experiment 1
\label{sec:exp1}
}

We validate our controller by testing a structure that has an $F$-frame that is not aligned with $\{S\}$. A single $R$-module forms such a structure, with all its rotors tilting 10 degrees in pitch. Thus, the structure of one module has the $F$-frame specified by ${}^S\boldsymbol{R}_F=\text{Rot}(y, \frac{\pi}{18})$. The structure tracks a 4-DOF vertical-helix trajectory in $\{W\}$. 
The helix centers at $(-0.5, 0)$ in the $xy$-plane with a radius of $0.45$ m and oscillates along the $z$-axis between $0.45$ m and $0.95$ m with a period of $14$ seconds. Meanwhile, the desired yaw angle increments with a period of 18 seconds. 
The trajectory tracking data are shown in Fig.~\ref{fig:exp1}. 
The average error in position is $\mu_x=0.0043m$, $\mu_y=-0.0045m$, and $\mu_z=0.0098m$, and the standard deviation is $\sigma_x=0.0398m$, $\sigma_y=0.0423m$, $\sigma_z=0.0437m$. The angular errors are $\mu_{\psi}=-0.4131^{\circ}$, and the standard deviation is $\sigma_{\psi}=3.6729^{\circ}$. 
The structure of one module has an attitude ${}^W\!\!\boldsymbol{R}_S={}^{S}\!\boldsymbol{R}^\top_F$ during hovering, which shows that the $F$-frame is indeed tracking the desired attitude.

\begin{figure}[t]
\centering
    \subfloat[\replytoallroundtwo{The trajectory tracking results of the structure in Experiment 1.
    \label{fig:exp1}}]{\includegraphics[width=0.43\linewidth]{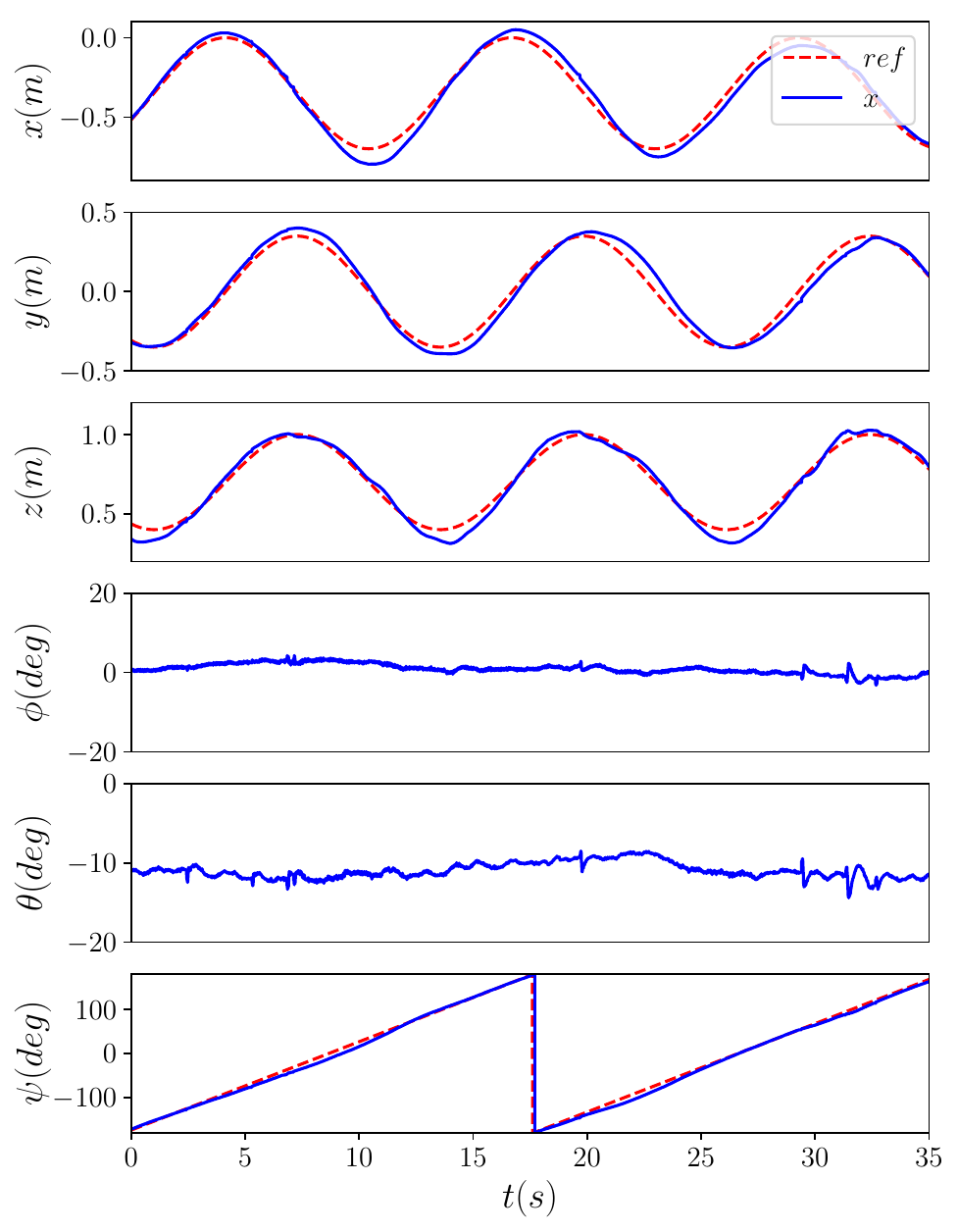}}\quad
    \subfloat[\replytoallroundtwo{The desired and recorded position and orientation of the structure in Experiment 2, tilted at $5^\circ$ versus $0^\circ$. 
    \label{fig:exp2}}]{
        \includegraphics[clip, trim=0.4cm 0.3cm 0.cm 0.cm, width=0.24\linewidth]{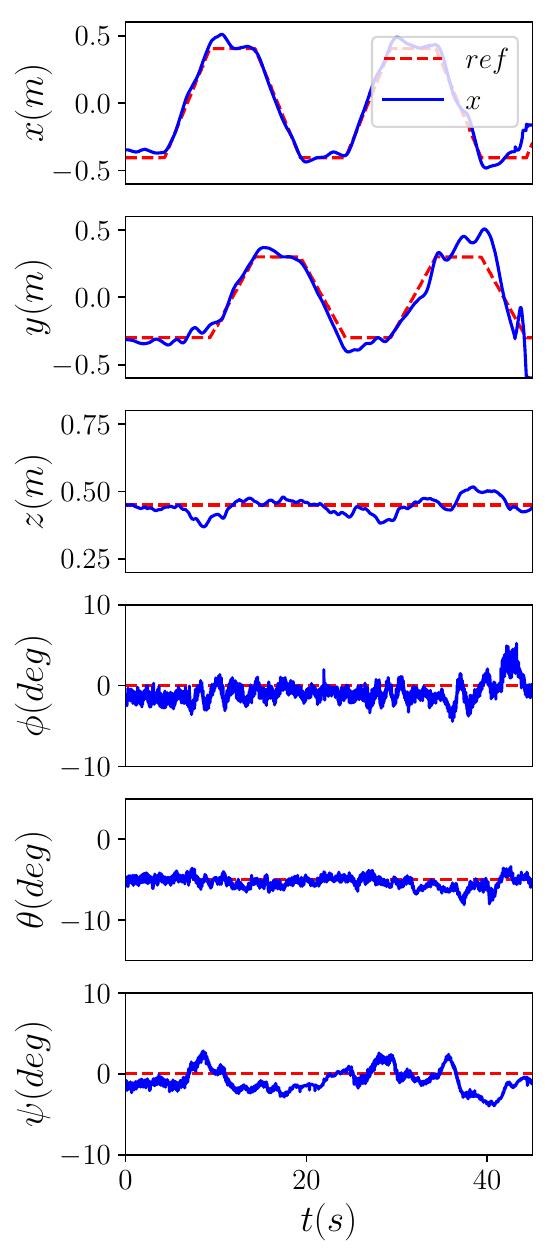}
        \includegraphics[clip, trim=0.4cm 0.3cm 0.cm 0.cm, width=0.24\linewidth]{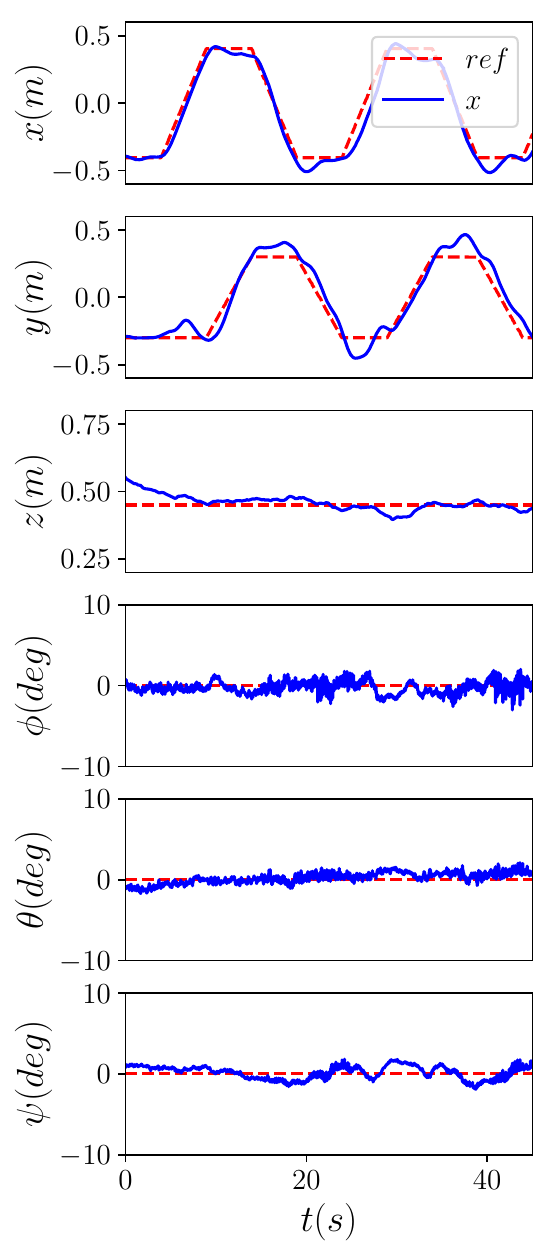}}\\
    \subfloat[\replytoallroundtwo{The trajectory tracking results of the structure in Experiment 3.
    \label{fig:exp3}}]{\includegraphics[width=0.46\linewidth]{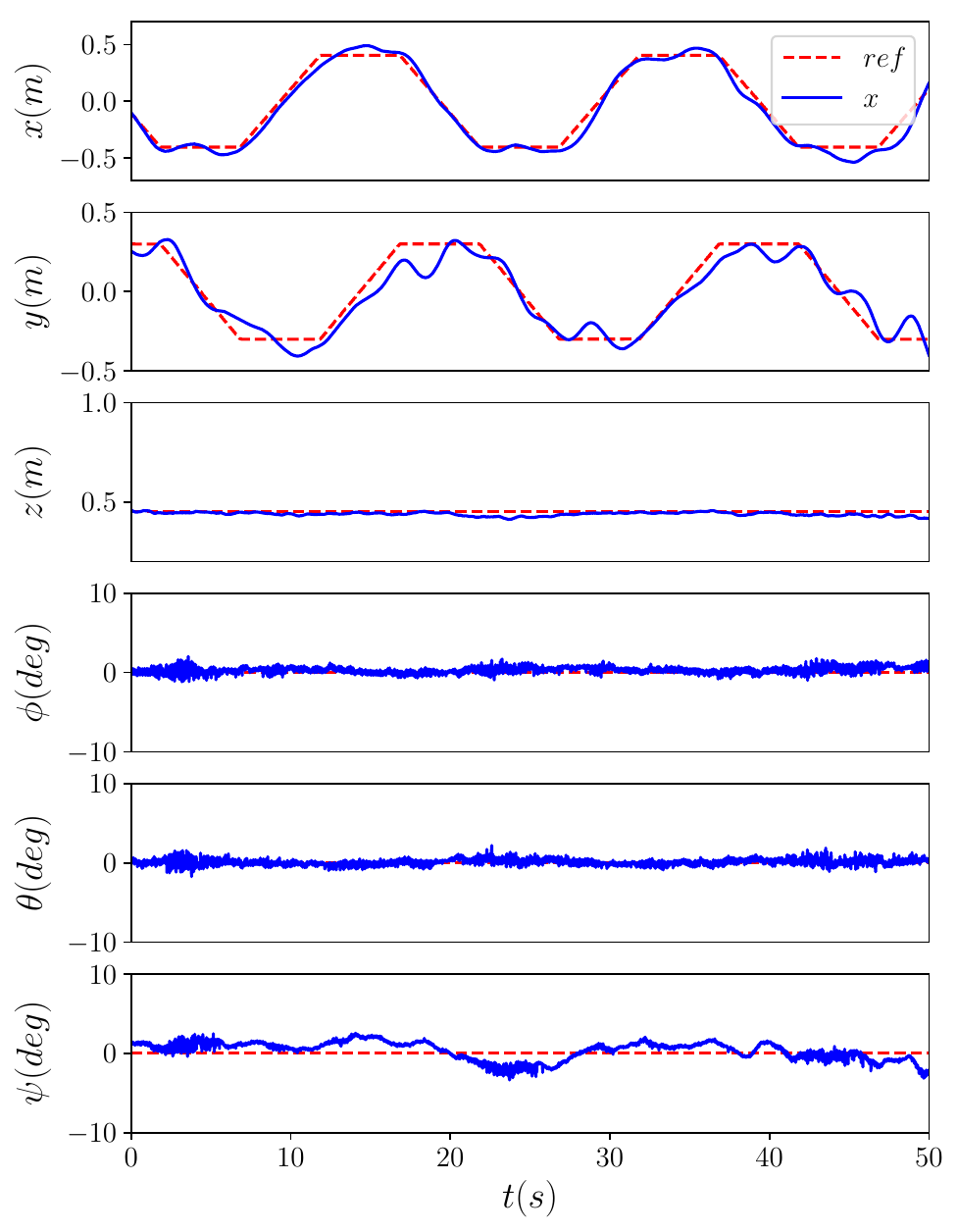}}\quad
    \subfloat[\replytoallroundtwo{The trajectory tracking results of the structure in Experiment 4.
    \label{fig:exp4}}]{\includegraphics[width=0.452\linewidth]{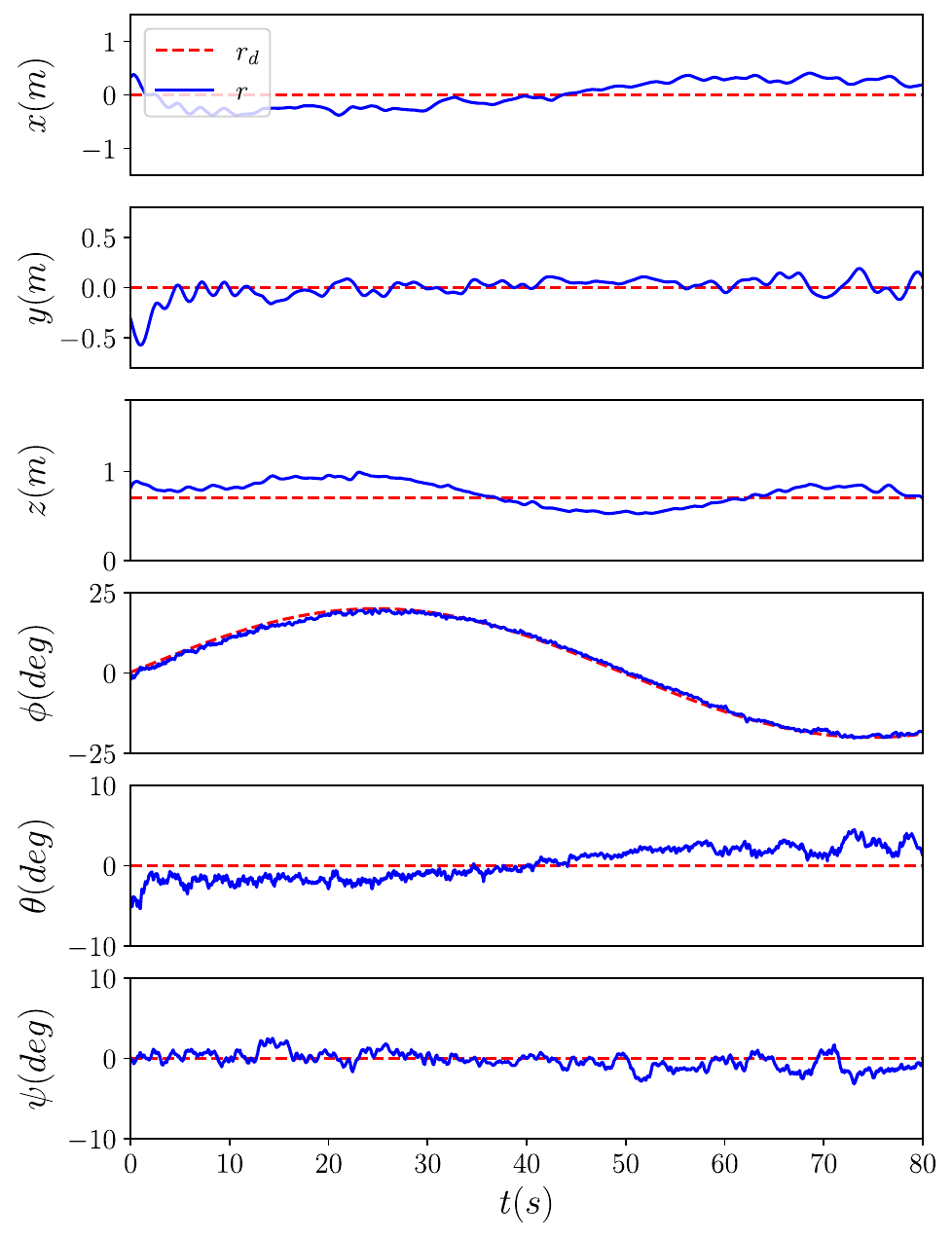}}\\
    \subfloat[\replytoallroundtwo{The trajectory tracking results of the structure in Experiment 5.
    \label{fig:exp5}}]{\includegraphics[width=0.45\linewidth]{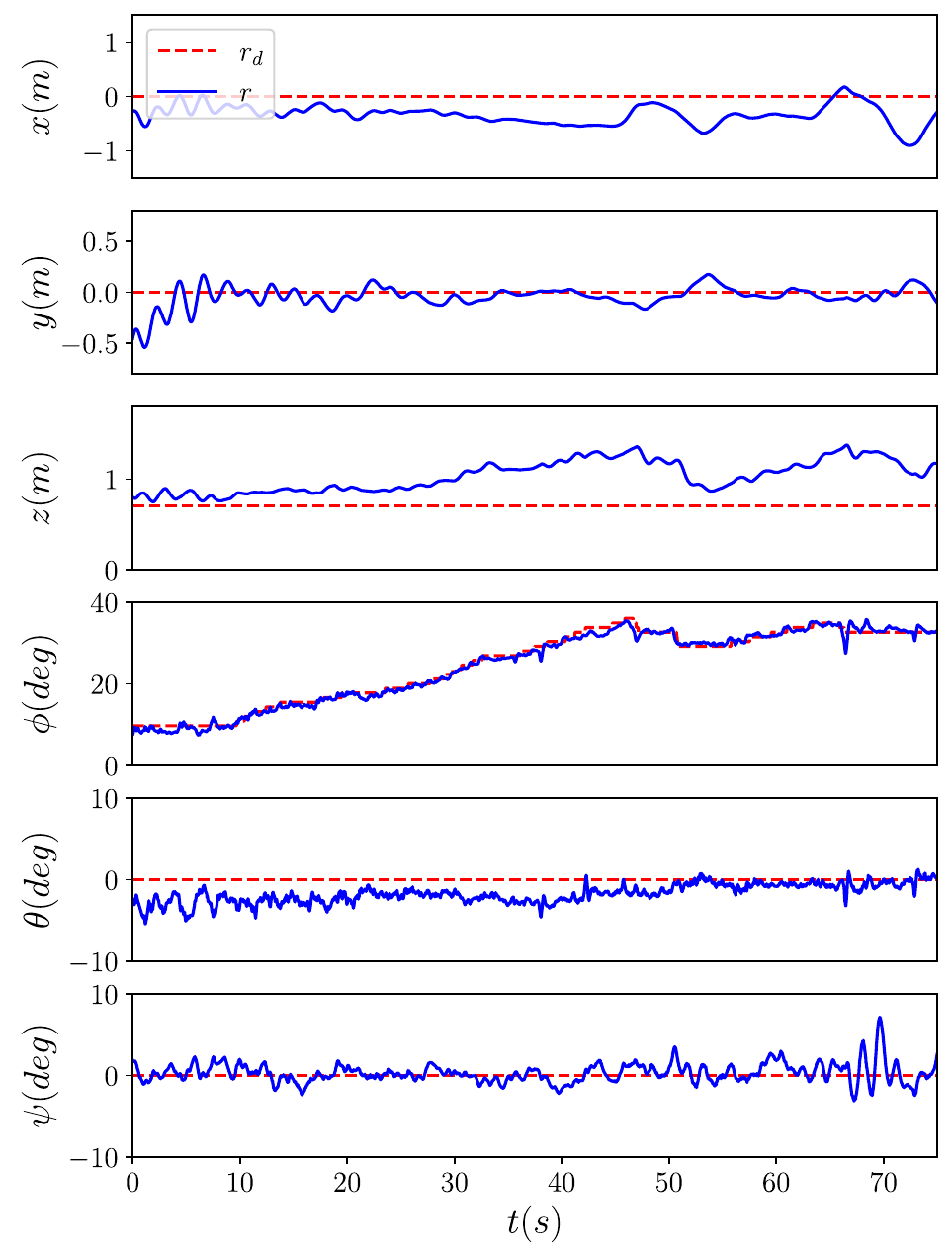}}\quad
    \subfloat[\replytoallroundtwo{The trajectory tracking results of the structure in Experiment 6.
    \label{fig:exp6}}]{\includegraphics[width=0.45\linewidth]{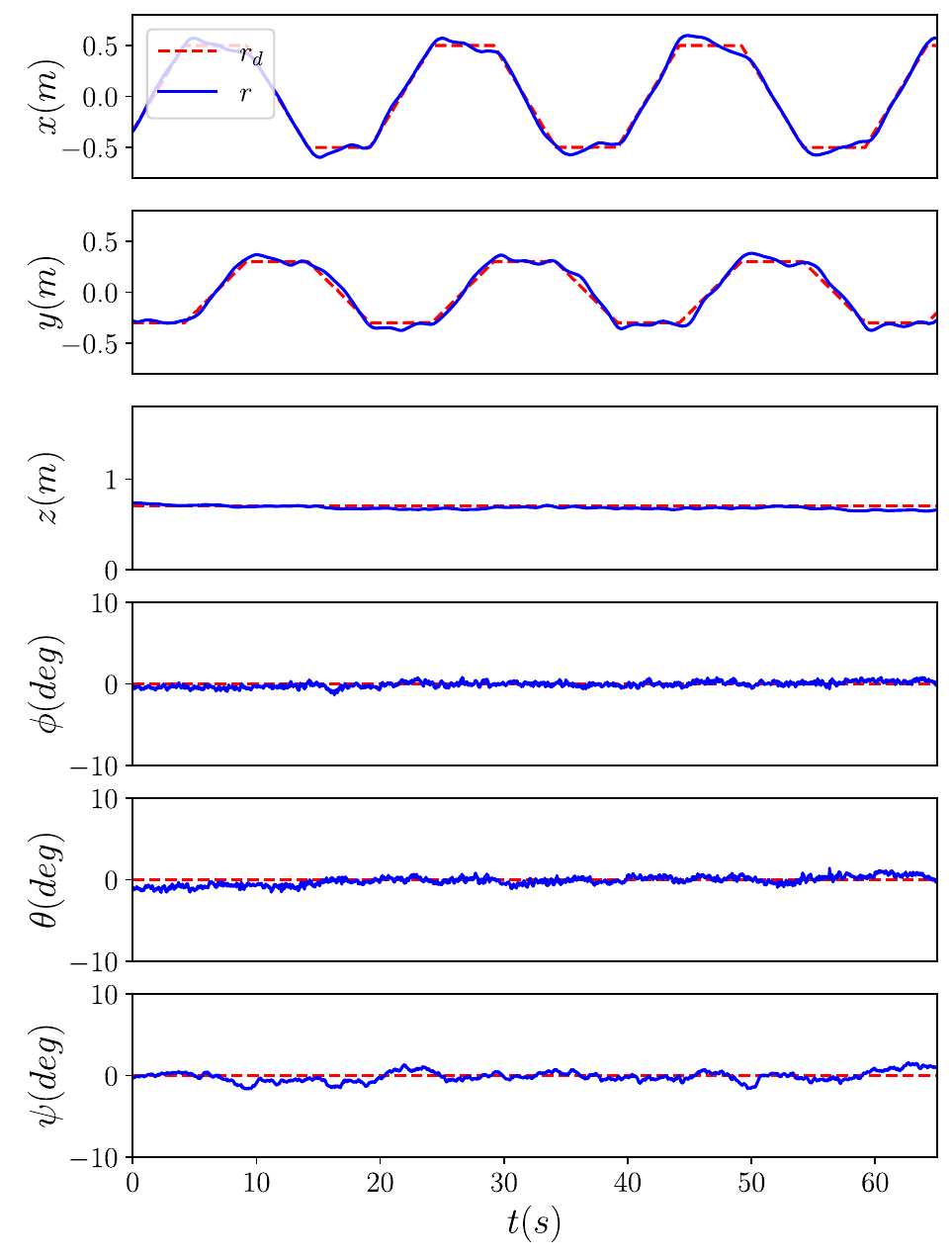}}
    \caption{\replytoallroundtwo{The state plots for trajectory tracking, in the order of $x$-, $y$-, $z$-translation and roll-, pitch-, and yaw-orientation, respectively. The red dashed curves represent the desired and the blue represent the recorded trajectories.}}
\end{figure}

\subsubsection{
Experiment 2
\label{sec:exp2}
}
We validate the \emph{dimensioning matrix} by testing a structure with 5 ADoF, composed of two $R$-modules. The structure follows a rectangular trajectory on the $xy$-plane of $\{W\}$ with a length of $0.8$ m and a width of $0.6$ m. One module has all its rotors tilting 30 degrees in pitch and the other tilting -30 degrees, resulting in 5 ADoF.
The resulted $F$-frame aligns with $\{S\}$, thus ${}^{S}\!\boldsymbol{R}_F = \boldsymbol{I}_3$. The structure tracks the trajectory when its $\{S\}$ remains a pitch angle of $0$ and $-5$ degrees, which shows the independence of its translation on the $x$-axis and its pitch. 
The average position error is $\mu_x=-0.0818m$, $\mu_y=-0.0059m$, and $\mu_z=0.0055m$, and the standard deviation is $\sigma_x=0.1227m$, $\sigma_y=0.0979m$, $\sigma_z=0.0322m$. The angular errors are $\mu_{\phi}=1.7718^{\circ}$, $\mu_{\theta}=1.3353^{\circ}$, $\mu_{\psi}=0.7125^{\circ}$, and the standard deviation is $\sigma_{\phi}=1.2356^{\circ}$, $\sigma_{\theta}=0.6866^{\circ}$, and $\sigma_{\psi}=1.4248^{\circ}$. 
The structure has the highest error along the $x$-axis despite the reference pitch angle, which is below $0.09$ m. 
Along the $y$- and $z$-axes, the structure keeps the error below $0.06$ m. 

\subsubsection{
\label{sec:exp3}
Experiment 3 - Rectangle tracking
}
We validate the control strategy for a fully-actuated structure. The structure is composed of four $R$-modules to track a $0.8\times0.6$ rectangular trajectory on the $xy$-plane of $\{W\}$ without any tilt. We assemble the modules in the structure as a $2\times2$ square in the $xy$-plane of $\{S\}$ and label the module in quadrants $2, 3, 4$ and $1$ of the $xy$-plane of $\{S\}$ as module $1, 2, 3,\text{ and }4$, respectively. In module $1, 3, 2,\text{ and }4$, the rotors are tilting 30, -30 degrees in pitch, -30, and 30 degrees in roll, respectively. Thus, ${}^{S}\!\boldsymbol{R}_{1} = {}^{S}\!\boldsymbol{R}^{\top}_{3} = \text{Rot}(y, \frac{\pi}{6})$ and ${}^{S}\!\boldsymbol{R}_{2} = {}^{S}\!\boldsymbol{R}^{\top}_{4} = \text{Rot}(x, -\frac{\pi}{6})$. The $F$-frame aligns with $\{S\}$ and ${}^{S}\!\boldsymbol{R}_F = \boldsymbol{I}_3$. The trajectory tracking data are shown in Fig.~\ref{fig:exp3}. The average error in position is $\mu_x=0.0039m$, $\mu_y=-0.0028m$, and $\mu_z=0.0207m$, and the standard deviation is $\sigma_x=0.0898m$, $\sigma_y=0.0783m$, $\sigma_z=0.0158m$. The angle error is $\mu_{\phi}=0.0609^{\circ}$, $\mu_{\theta}=0.0306^{\circ}$, $\mu_{\psi}=0.1411^{\circ}$, and the standard deviation is $\sigma_{\phi}=0.4735^{\circ}$, $\sigma_{\theta}=0.3255^{\circ}$, and $\sigma_{\psi}=0.6361^{\circ}$. 
Along the $y$-axis, the structure has the largest error, but below $0.15$ m. Along the $x$- and $z$-axes, the structure keeps the error under $0.05$ m. When the structure is tracking the trajectory, both roll and pitch angles remain 0 degrees, which shows the independence of translation and rotation.

\subsubsection{
\label{sec:exp4}
Experiment 4
}
\replytoall{In initial tests not included in this manuscript, we observed that the H-ModQuad configuration in \replytotworoundtwo{Experiment 3} has very limited tilting capability due to unidirectional motor constraints. Commanding the structure to tilt as little as $5^\circ$ often resulted in crashes, which motivated the design of a translation-only trajectory for \replytotworoundtwo{Experiment 3} to demonstrate independent rotation and translation.} Therefore, we assemble a structure composed of four $T$-modules, showing that the 6-ADoF structure can tilt without translation. The four $T$-modules are of two types, one with $\eta=\frac{\pi}{4}$ and the other with $\eta=-\frac{\pi}{4}$. The two types of $T$-modules are placed diagonally as a $2\times2$ square in the $xy$-plane of $\{S\}$, as shown in Fig.~\ref{fig:HModQuads}(c), and ${}^{S}\!\boldsymbol{R}_F = \boldsymbol{I}_3$. We command the structure to track a sinusoidal curve in pitch while hovering in place. The curve has a period of 90 seconds and \replytoall{a magnitude of $20^\circ$}. Because the structure is centrosymmetric, the capability to tilt in roll and pitch is equivalent. The trajectory tracking data are shown in Fig.~\ref{fig:exp4}. 
The average error in position is $\mu_x= -0.0104m$ , $\mu_y=-0.0143m$, and $\mu_z=-0.0555m$, and the standard deviation is $\sigma_x=0.2453m$, $\sigma_y=0.0687m$, $\sigma_z=0.1318m$. The angle error is $\mu_{\phi}=-0.3988^{\circ}$, $\mu_{\theta}=-0.6541^{\circ}$, $\mu_{\psi}=0.3093^{\circ}$, and the standard deviation is $\sigma_{\phi}=1.8583^{\circ}$, $\sigma_{\theta}=2.435^{\circ}$, and $\sigma_{\psi}=1.019^{\circ}$. 
The structure keeps the position error below $0.3m$ and the rotation error under $2.5^{\circ}$, changing roll while hovering. The position error is mainly caused by the mismatch of the prototype with the \emph{Motor Dynamics} model, which is compensated by the controller.

\subsubsection{
\label{sec:exp5}
Experiment 5
}
Experiment 4 shows that the structure in Fig.~\ref{fig:HModQuads}(c) has a better tilting ability than that in \replytotworoundtwo{Experiment 3}. In order to find the maximum tilt angle of this structure with four $T$-modules, we design another experiment to manually increase the reference roll until the structure is unable to hover in place while tracking the tilt angle. The trajectory tracking data are shown in Fig.~\ref{fig:exp5}. We observe that as the reference tilt angle increases, some rotors stop rotating due to saturation, as shown in Fig.~\ref{fig:titlePic}. When the commanded rotor force is below a threshold, the linear increase of the thrust force to the PWM no longer holds, leading to a mismatch between the theoretical model and the physical system of the H-ModQuad. 
The average error in position is $\mu_x=0.3389m$, $\mu_y=0.0309m$, and $\mu_z=-0.3451m$, and its standard deviation is $\sigma_x=0.1898m$, $\sigma_y=0.0941m$, $\sigma_z=0.1659m$. The angle error is $\mu_{\phi}=1.5495^{\circ}$, $\mu_{\theta}=0.3805^{\circ}$, $\mu_{\psi}=-0.2988^{\circ}$, and their standard deviation is $\sigma_{\phi}=1.1736^{\circ}$, $\sigma_{\theta}=0.971^{\circ}$, and $\sigma_{\psi}=1.2678^{\circ}$. 
Along the $x$-axis, the error reaches $0.8m$ when the structure is approaching the maximum tilt. \replytoall{Based on this experiment, the robot achieves a maximum roll angle of $38.0^\circ$, which aligns with the actuation polytope analysis.}

\subsubsection{
\label{sec:exp6}
Experiment 6
}
In this experiment, we have a task similar to that of \replytotworoundtwo{Experiment 3} but using a structure composed of $T$-modules. The trajectory tracking data are shown in Fig.~\ref{fig:exp6}. 
The average error in position is $\mu_x=0.0039m$, $\mu_y=-0.0028m$, and $\mu_z=0.0207m$, and its standard deviations $\sigma_x=0.0898m$, $\sigma_y=0.0783m$, $\sigma_z=0.0158m$. The angle error is $\mu_{\phi}=0.0609^{\circ}$, $\mu_{\theta}=0.0306^{\circ}$, $\mu_{\psi}=0.1411^{\circ}$, and their standard deviation $\sigma_{\phi}=0.4735^{\circ}$, $\sigma_{\theta}=0.3255^{\circ}$, and $\sigma_{\psi}=0.6361^{\circ}$. 
The positional error is below $0.04$ m along all axes, and the rotational error is below $0.5$ degrees along all axes. Combining Experiments 4 and 6, we show in full that this H-ModQuad of 4 $T$-modules is fully actuated and has control over 6 DOF.

\begin{figure}[t]
\centering
    \subfloat[\replytotworoundtwo{ 
    The position errors in trajectory tracking across the six experiments.
    \label{fig:pos_error}}]{\includegraphics[trim={0 3mm 0 0},clip,width=\linewidth]{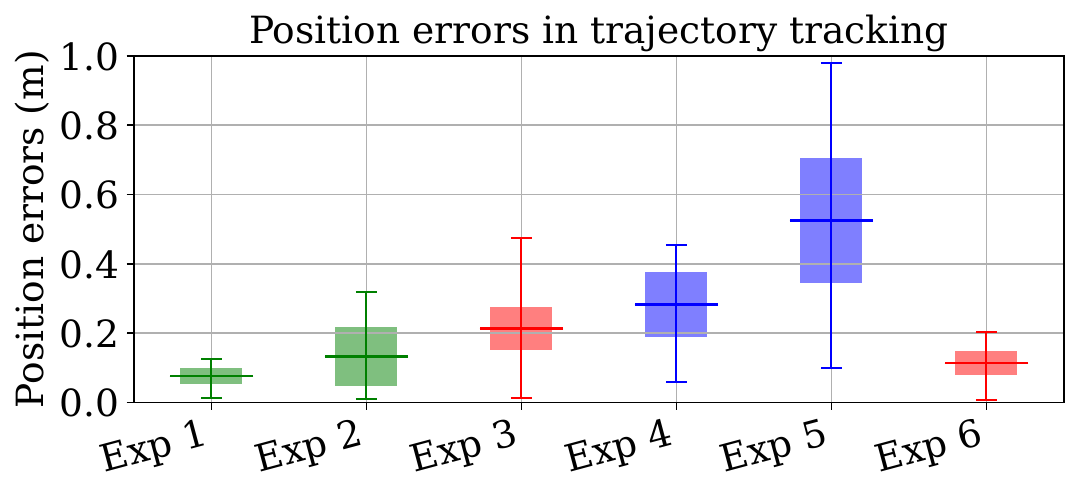}}\\
    \subfloat[\replytotworoundtwo{
    The orientation errors in trajectory tracking across the six experiments.
    \label{fig:ori_error}}]{\includegraphics[trim={0 3mm 0 0},clip,width=\linewidth]{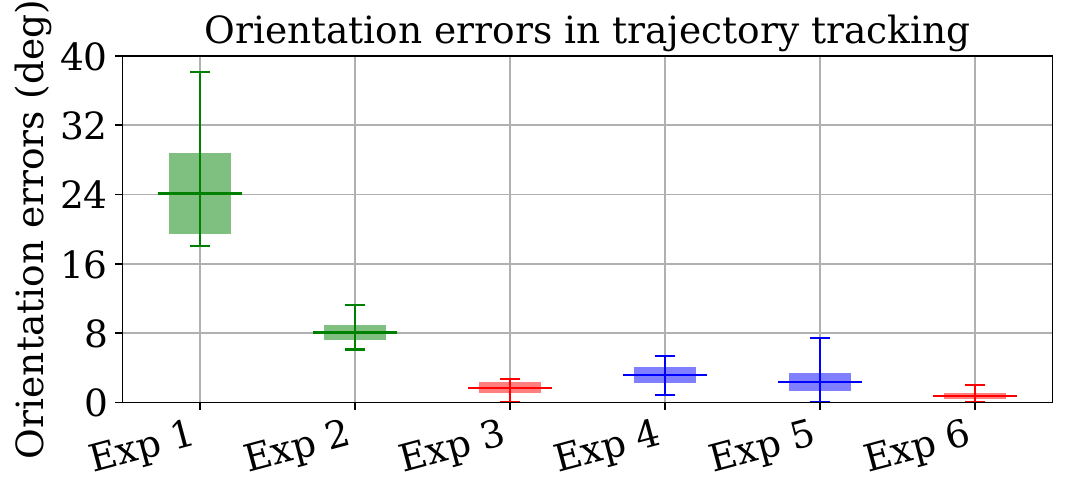}}
    \caption{\replytotworoundtwo{The error statistics of the six experiments. The green T-string entries are associated with the experiments using the structures of 4 and 5 ADoF, composed of $R$-modules. The blue entries are associated with the orientation-tracking experiments using the structure of 6 ADoF, composed of 4 $T$-modules. The red entries are associated with the performance comparison between the fully-actuated structure composed of 4 $R$-modules (Exp 3) and the one of 4 $T$-modules (Exp 6).}}
\end{figure}

\subsection{Discussion}
The experiments with the $R$-modules show that the controller applies effectively despite the ADoF of the structure. We note that although having 6 ADoF when assembling four $R$-modules together, \replytoone{the structure in \replytotworoundtwo{Experiment 3} loses stability during initial tests when attempting to track a varying tilting angle independently from translation, which we opted not to include in the manuscript.} The observation matches the maximum tilt of the structure obtained from the force polytope, which is $12^\circ$. \replytotworoundtwo{It also infers that the theoretical maximum force that the structure can generate in the horizontal plane is less than $\tan{12^\circ}\approx0.21$ times that in the $z$-axis, which limits its translation. As shown in Fig.~\ref{fig:pos_error}, the structure experiences a large positional error, which is consistent with our analysis.}

The fully-actuated structure of $T$-modules in Experiment~4-6 can push the tilt angle over $35^\circ$ without losing stability, as shown in the blue T-string plots in Fig.~\ref{fig:ori_error}, even if some motors saturate, which is not achievable using the structure composed of only $R$-modules. This is highlighted in Experiment 5. Compared to Experiment 3 where the structure composed of 4 $R$-modules are following the same trajectory as in Experiment 6, the structure composed of 4 $T$-modules achieves significantly better position tracking performance with a mean error of $0.1138m$ and standard deviation of $0.0347m$, over the mean error of $0.2134m$ and standard deviation of $0.06117m$ achieved in Experiment 3, as shown in the red T-string plots of Fig.~\ref{fig:pos_error}, highlighting the greater actuation capabilities brought by the newly introduced $T$-modules.
However, we also notice an issue related to the model of the structure. In Experiments 4 and 5, the attitude change affects the quality of position tracking, as shown in Fig.~\ref{fig:pos_error}, especially on the $z$-axis. Despite the stability, position tracking drifts along all three axes as the tilt angle increases. The drift is due to the mismatch between the modeling and the physical prototype. In \eqref{eq:Acomponents}, we show that the design matrix $\boldsymbol{A}$ depends on the positions of the motors, which is not directly measurable and has greater impact on a $T$-module. The thrust of rotors applies directly at the position of the motor mounts in our model, which is different from the actual force as a result of the aerodynamic properties of the rotors. We mitigate this problem by applying an integral term in the controller. 

\section{Conclusion}
We present a modular UAV that can increase its strength and ADoF from 4 to 5 and 6 with two torque-balanced module designs, $R$- and $T$-module. The module configuration determines the ADoF. We model the actuation capabilities of the structures using the actuation ellipsoid and actuation polytope. The polytope is an analysis tool to check whether a structure can satisfy certain tasks requirements such as hovering at a certain tilt angle.
We use the actuation ellipsoid to find the best frame for a structure. 

In future work, we would like to explore more configurations of H-ModQuad such as omnidirectionality. Also, since different configurations satisfy different task requirements, we want to study systematic approaches to generate suitable designs given task requirements and available modules.

\section{Acknowledgments}
The authors gratefully acknowledge the support of the NSF Awards 2322840, 2442475, and ONR Award 544835.


\bibliographystyle{IEEEtran}
\bibliography{ref}

\begin{IEEEbiography}[{\includegraphics[width=1in,height=1.25in,clip,keepaspectratio]{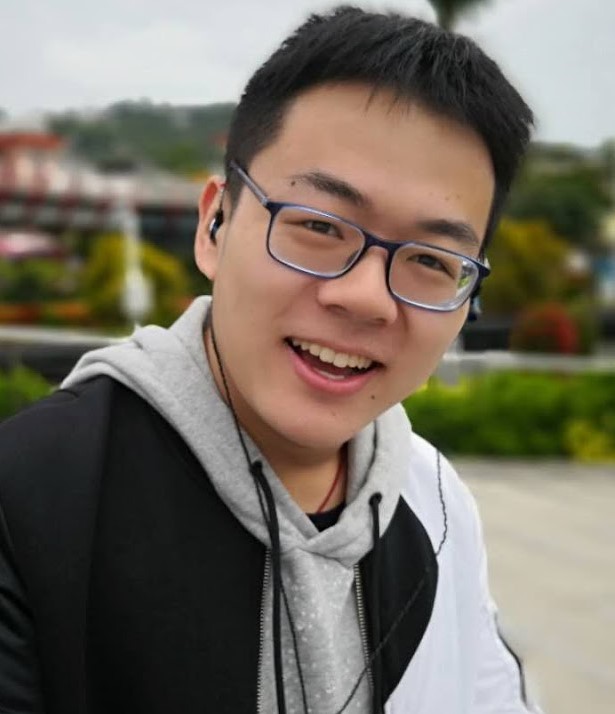}}]{Jiawei Xu} is a doctoral candidate in Computer Science and Engineering, and a member of Swarms Lab and Autonomous and Intelligent Robotics Laboratory (AIRLab) at Lehigh University, Bethlehem, PA. He worked as a visiting scholar at The University of New Mexico. He received his B.Sc. (2019) in Electrical Engineering and Computer Systems Engineering from Rensselaer Polytechnic Institute.
His current research focuses on multirotor aerial vehicles, modular robots, aerial physical interactions, adaptive, optimal, and learning-based control for aerial manipulation.
\end{IEEEbiography}
\begin{IEEEbiography}[{\includegraphics[width=1in,height=1.25in,trim=8cm 13cm 8cm 0cm,clip,keepaspectratio]{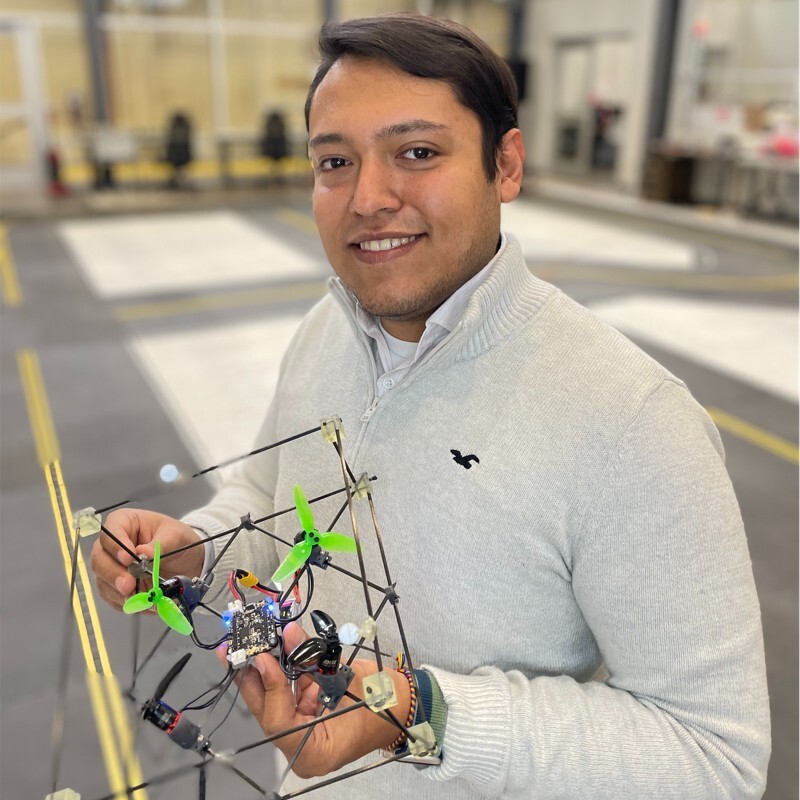}}]{Diego S. D'Antonio} (S'10-M'15) is a doctoral candidate in Computer Science and Engineering at Lehigh University, Bethlehem, PA, USA. He is affiliated with the Swarms Lab and Autonomous and Intelligent Robotics Laboratory (AIRLab) at Lehigh University. He received his M.Sc. in Control Engineering from the University of Ibagué, Colombia.
From 2012 to 2015, he served as an engineering and innovation project manager at Ideas Disruptivas, Mexico. He also held positions as the chair of the IEEE student branch chapter and chair of the Industrial Application Society at the University of Ibagué from 2017 to 2019.
His current research focuses on multi-rotor systems, cooperative flying robots, and aerial cable manipulation, with a particular emphasis on using cables for object interaction and transportation.
\end{IEEEbiography}
\begin{IEEEbiography}[{\includegraphics[width=1in,height=1.25in,clip,keepaspectratio]{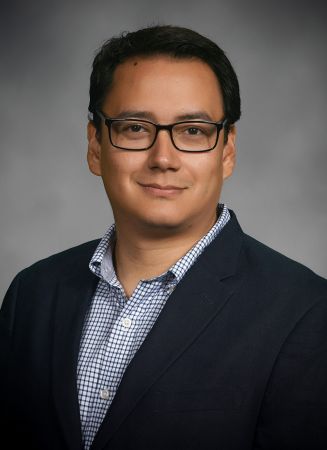}}]{David Salda\~na} is an Assistant Professor in Computer Science and Engineering, and director of the Swarms Lab at Lehigh University. 
His research focuses on multi-robot systems, swarm robotics, and aerial manipulation, with applications in environmental monitoring, disaster response, and construction. Saldaña’s work aims to develop resilient, adaptive robotic systems capable of operating in dynamic and unpredictable environments.

He worked as a Post-Doctoral Researcher at the GRASP Laboratory at University of Pennsylvania.
He earned his PhD in Computer Science, artificial intelligence and robotics, from the Federal University of Minas Gerais in Brazil in 2017, and holds both M.Sc. and B.Sc. degrees in systems engineering from the Universidad Nacional de Colombia. His research has been recognized with support from the National Science Foundation and the Office of Naval Research, including the NSF CAREER award.
\end{IEEEbiography}

\end{document}